%% file: neurips_2026.tex
\documentclass{article}

\PassOptionsToPackage{numbers, compress}{natbib}
\usepackage[preprint]{neurips_2026}

\usepackage[utf8]{inputenc} 
\usepackage[T1]{fontenc}    
\usepackage{url}            
\usepackage{booktabs}       
\usepackage{amsfonts}       
\usepackage{nicefrac}       
\usepackage{microtype}      
\usepackage[table,xcdraw,dvipsnames]{xcolor}
\usepackage[
    colorlinks=true,
    linkcolor=Maroon,    
    citecolor=RoyalBlue, 
    urlcolor=magenta     
]{hyperref}
\usepackage{graphicx}
\usepackage{wrapfig}
\usepackage{amsmath}
\usepackage{multirow}
\usepackage{pifont}
\usepackage{caption}
\usepackage[table,xcdraw]{xcolor}
\usepackage[normalem]{ulem}
\usepackage[most]{tcolorbox}
\usepackage{xcolor}
\usepackage{booktabs}
\usepackage{adjustbox}

\definecolor{casegray}{HTML}{4A4A4A}
\definecolor{caseblue}{HTML}{3F6FA8}
\definecolor{casered}{HTML}{A94442}
\definecolor{casegreen}{HTML}{1B8A5A}
\definecolor{lightblue}{HTML}{F3F8FF}
\definecolor{lightred}{HTML}{FFF5F3}
\useunder{\uline}{\ul}{}

\newcommand{\wrong}[1]{\textcolor{red!75!black}{\textbf{#1}}}
\newcommand{\correct}[1]{\textcolor{green!45!black}{\textbf{#1}}}
 
\title{DynEdit: Dynamic Entropy-Guided Sequential Editing for Large Language Models}

\author{
 \textbf{Jinhu Fu\textsuperscript{1}}\thanks{fjhu@bupt.edu.cn. Work done during internship in DreamX Team.},
 \textbf{Yan Bai\textsuperscript{2}},
 \textbf{Yihang Lou\textsuperscript{2}},
 \textbf{Li Sun\textsuperscript{1}},
 \textbf{Jiahong Wu\textsuperscript{3}}\thanks{Project Leader.},
\\
 \textbf{Xiangxiang Chu\textsuperscript{3}},
 \textbf{Sen Su\textsuperscript{1,4}}\thanks{Corresponding author.},
\\
 \textsuperscript{1}Beijing University of Posts and Telecommunications,
 \textsuperscript{2}Peking University,
\\
 \textsuperscript{3}DreamX Team,
 \textsuperscript{4}Chongqing University of Posts and Telecommunications
}

\begin{document}

\maketitle

\begin{abstract}
\label{abs}
Knowledge editing aims to update outdated or incorrect knowledge in large language models without full retraining. Existing parameter-editing methods remain limited for long-sequence knowledge: single-point editors provide insufficient coverage for extended targets, while fixed chunk-based editors distribute editing effort uniformly across rigid predefined windows, failing to prioritize difficult, information-dense tokens. In this work, we show that token entropy is closely associated with both editing difficulty and semantic importance: high-entropy regions are harder to edit and often contain key factual content. Motivated by this observation, we propose DynEdit, a dynamic framework for long-sequence knowledge editing. DynEdit uses token-level entropy to guide non-uniform sliding, increasing the coverage of difficult high-entropy regions, and further applies entropy-guided token-level optimization to strengthen updates on information-dense tokens within each editing window. Experiments on three benchmarks and two backbone models show that DynEdit consistently outperforms competitive baselines across diverse long-form editing settings, achieving gains of up to +14.0 BLEU points on challenging paraphrase queries and near-saturated semantic scores on several diverse knowledge domains.

\end{abstract}

\section{Introduction}

\label{intro}


Large language models (LLMs) have achieved remarkable progress across a wide range of tasks \cite{vaswani2017attention, brown2020language, liu2024tuning, chen2024evlm}. However, they inevitably encode outdated, incorrect, or conflicting factual knowledge in their parameters \cite{huangtransformer, brown2020language, geva2021transformer}. Knowledge editing addresses this limitation by modifying model behavior toward target facts without requiring full retraining \cite{de2021editing, mitchellfast}. This enables efficient and targeted knowledge updates while preserving the model’s general capabilities \cite{madaan2022memory, roberts2020much, zhang2024locate}.

Existing knowledge editing methods can be broadly grouped into two paradigms \cite{zhang2024comprehensive, wang2024wise, yu2024melo}. Parameter-editing methods, such as MEMIT \cite{meng2022mass}, AlphaEdit \cite{fang2024alphaedit} and AnyEdit \cite{2025arXiv250205628J}, directly update model parameters by locating and modifying knowledge-relevant representations. Context-augmentation methods, such as IKE \cite{zheng2023can}, EditCoT \cite{EditCoT} and LTE \cite{jiang-etal-2024-learning}, instead inject edited knowledge through prompts or external contexts, making them suitable for black-box settings \cite{fang2024alphaedit}. This work focuses on the parameter-modifying paradigm, where is to directly update and fix factual knowledge stored in model parameters.

Despite substantial progress, most parameter-modifying methods are still designed around localized editing \cite{meng2022locating}. Methods such as MEMIT \cite{meng2022mass}, AlphaEdit \cite{fang2024alphaedit} and so on \cite{fu2026learning, fu2025knowledge},  typically compute updates from a small number of fact-relevant positions, often centered on the subject representation or the final token of the subject. This design is effective for triplet-style factual edits, where the target knowledge is short and structurally compact. However, it becomes inadequate for long-sequence knowledge editing \cite{su2025mu, 2025arXiv250205628J}. When the target update is an extended free-form answer, a single localized editing position provides limited supervision coverage and cannot adequately capture the dependencies distributed across the whole sequence. We refer to this limitation as the \emph{single-token editing bottleneck}.

To mitigate this bottleneck, recent methods such as UnKE \cite{deng2024everything} and AnyEdit \cite{2025arXiv250205628J} extend editing from a single position to multiple target positions. UnKE optimizes edits over long target sequences, while AnyEdit introduces an autoregressive sliding-window strategy that updates the last position within each window. These approaches substantially improve the coverage of long-form knowledge editing. Nevertheless, they still suffer from two critical limitations:

\begin{figure}[h]
  \centering
  \includegraphics[width=\textwidth]{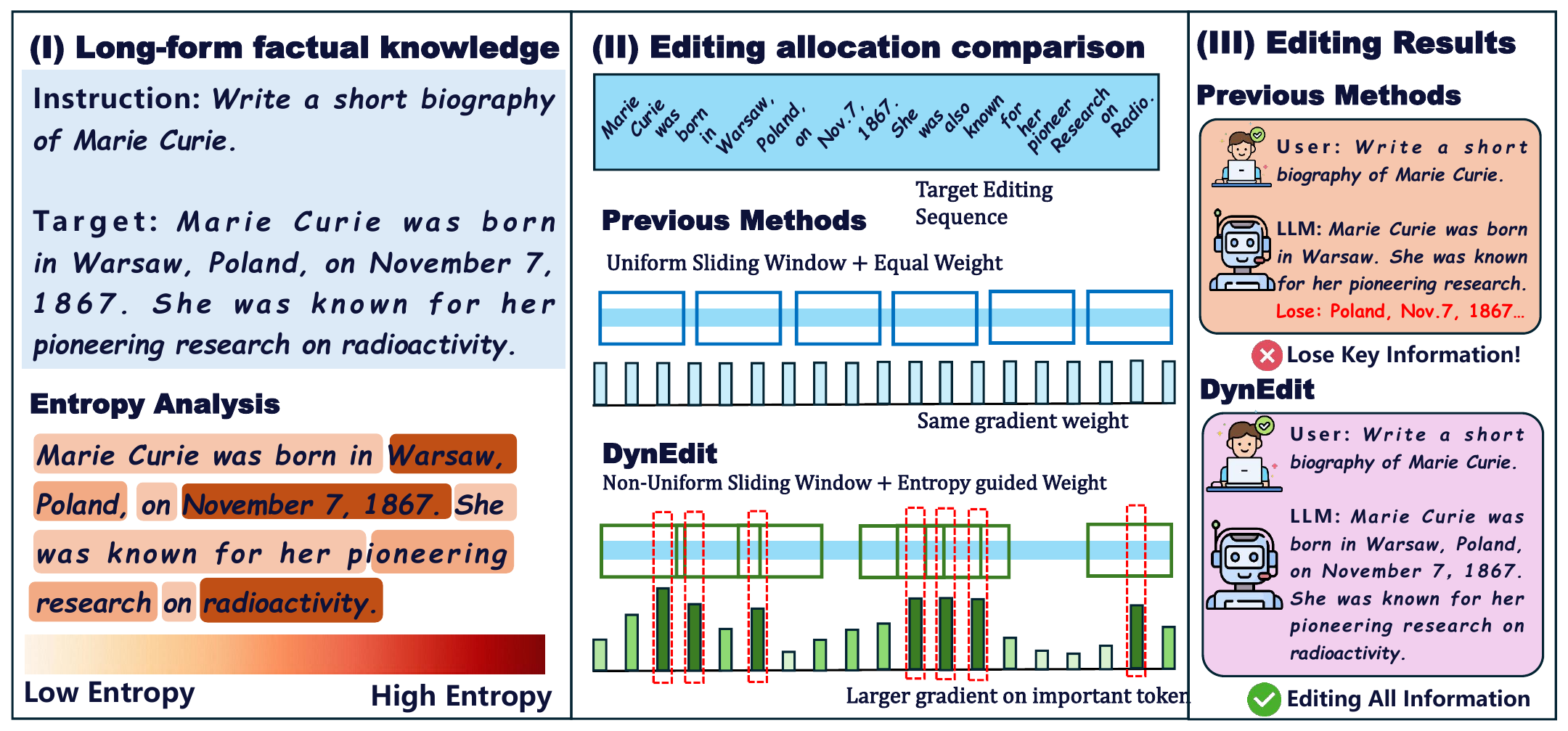}
  \caption{Revealing the deficiencies in existing knowledge editing approaches.}\label{fig:1}
\vspace{-3mm}
\end{figure}

\ding{224} \textbf{Difficulty-agnostic editing allocation.}
As illustrated in Figure~\ref{fig:1}, long-form target knowledge exhibits highly non-uniform editing difficulty: some regions are easy and predictable, while others contain high-uncertainty factual details that are harder to rewrite correctly. However, existing sequence-level editors allocate editing effort in a largely difficulty-agnostic manner. UnKE optimizes over target tokens without explicitly distinguishing easy and difficult regions, while AnyEdit adopts a fixed sliding-window strategy with a predefined window size and stride. Such static allocation inevitably spends redundant computation on low-entropy regions and provides insufficient editing density for high-entropy regions, where key factual information is often concentrated \cite{xiong2025hs, wang2025position}.

\ding{224} \textbf{Token-Level Optimization Imbalance under Fixed Chunking.} Existing long-sequence editors usually divide the target sequence into fixed chunks and update the editing vector at the last position of each chunk \cite{su2025mu, 2025arXiv250205628J}. Under this design, tokens within the same chunk are typically optimized with comparable strength, regardless of their roles in determining the factual content \cite{dai2026harder}. As a result, the optimization signal can be diluted by function words and other low-information tokens, while key factual tokens such as entities, dates, and locations do not receive sufficiently concentrated supervision \cite{liu2025token, fadeeva2024fact}. As illustrated in Figure~\ref{fig:1}, previous methods may therefore preserve the general form of the answer but still miss critical details.

To address the aforementioned challenges, we propose \textbf{DynEdit}, a dynamic entropy-guided sequence editing framework for large language models. DynEdit is built on the observation that long-form knowledge is highly non-uniform: high-entropy tokens are often both harder to edit and more semantically informative \cite{dong2025agentic, wang2025beyond}. Based on this insight, DynEdit allocates editing capacity according to token-level uncertainty through two coordinated designs.

\textbf{(I) Entropy-Guided Non-Uniform Sliding.}
DynEdit first estimates the token-level entropy \cite{dong2025agentic} of the target sequence under the pre-edit model, and uses it to adaptively control the sliding stride. In low-entropy regions, DynEdit adopts a larger stride to pass through predictable content efficiently. In high-entropy regions, it reduces the stride and increases window overlap, allowing difficult factual segments to be covered by more editing steps. This design improves the editing exposure of informative tokens, instead of relying on a fixed chunking schedule.

\textbf{(II) Entropy-Guided Token-Level Optimization.}
Beyond increasing the coverage of high-entropy regions, DynEdit further differentiates the optimization strength of individual tokens. Specifically, it assigns larger optimization weights to high-entropy tokens \cite{liu2025token} within each editing window, so that entities, dates, locations, and other information-dense tokens receive stronger supervision than low-information function words. This token-level objective complements the non-uniform sliding strategy: dense sliding increases how often difficult regions are edited, while entropy-guided weighting increases how strongly critical tokens are optimized once they are covered.

Experimental results demonstrate the effectiveness of DynEdit from three representative aspects:
\ding{182} \textbf{Stronger long-sequence editing.} DynEdit consistently improves both MEMIT- and UnKE-based editors, achieving up to \textbf{+18.96 BLEU} \cite{papineni-etal-2002-bleu} over AnyEdit$^\ast$ on UnKE-Paraphrase. \ding{183} \textbf{Robustness across formats and lengths.} On EditEverything \cite{2025arXiv250205628J}, DynEdit shows strong gains on structurally complex knowledge, e.g., \textbf{+27.03 Rouge-L} \cite{lin2004rouge} on math; length-bucket analysis further shows a \textbf{+23.88 BLEU} gain on the longest sequences. \ding{184} \textbf{Preserved general ability.} DynEdit maintains comparable general ability on \textbf{MMLU} \cite{hendrycks2020measuring} and \textbf{IFEval} \cite{zhou2023instruction}, while DynEdit$^\ast$ achieves the best results on most general benchmarks under the UnKE backbone.

Our contributions are summarized as follows:
\begin{itemize}
    \item We identify the non-uniform nature of long-sequence knowledge editing and show that high-entropy tokens are often both harder to edit and more semantically informative. Based on this observation, we introduce entropy as a principled signal for allocating editing capacity in long-form knowledge editing.
    \item We propose DynEdit, a dynamic entropy-guided editing framework that combines non-uniform sliding with token-level weighted optimization. The former increases the editing exposure of difficult high-entropy regions, while the latter strengthens the optimization contribution of informative tokens within each editing window.
    \item Extensive experiments on UnKE, CounterFact \cite{meng2022locating}, MQuAKE \cite{zhong2023mquake}, and EditEverything across Llama-3-8B \cite{grattafiori2024llama} and Qwen-2.5-7B \cite{qwen2_5_technical_report} and Qwen-3-8B \cite{yang2025qwen3} demonstrate that DynEdit consistently improves long sequence capability of diverse editors. Further analyses on ablation, diverse-formatted knowledge, general ability, and efficiency validate the effectiveness and robustness of our design.
\end{itemize}

\section{Motivation}
\label{sec:motivation}

In this section, we present preliminary analyses to examine the role of token entropy in long-sequence knowledge editing. These analyses show that entropy is closely associated with both editing difficulty and semantic importance, thereby motivating the design of DynEdit.

\begin{figure}[h]
  \centering
  \includegraphics[width=0.8\textwidth]{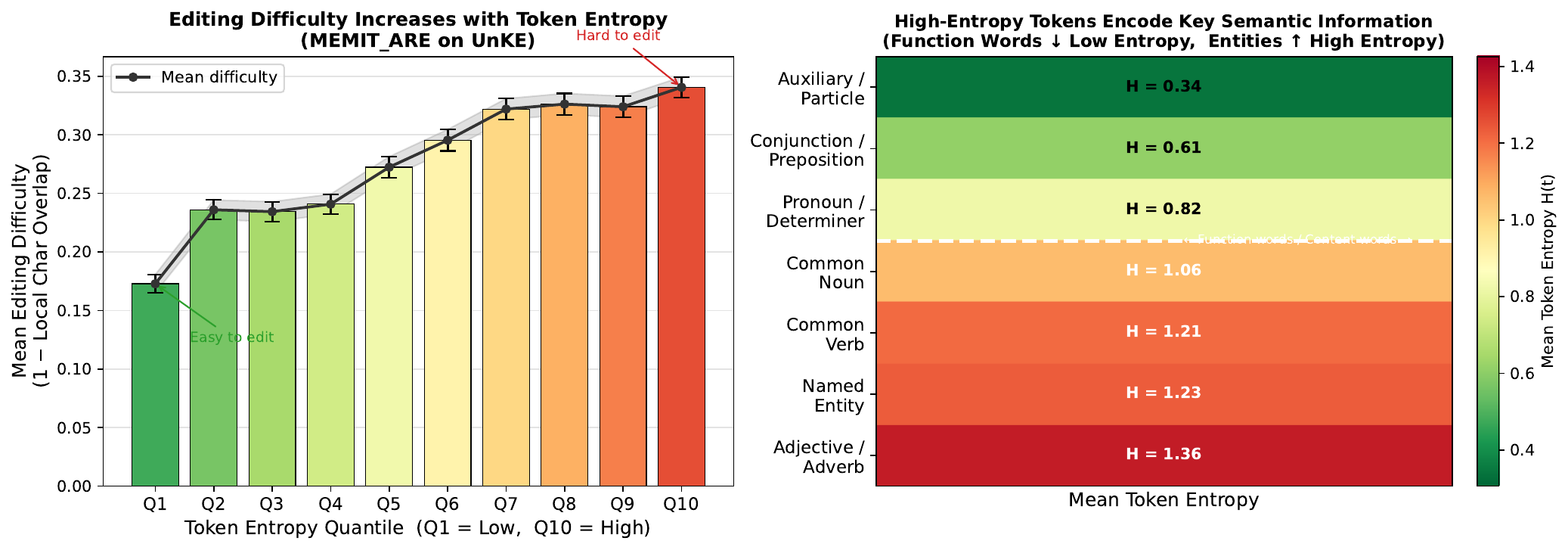}
  \caption{Token-level analysis of entropy, editing difficulty, and semantic role. High-entropy tokens are both harder to edit and more likely to carry core semantic information.}
  \label{fig:2}
  \vspace{-3mm}
\end{figure}

According to Figure~\ref{fig:2}, the left panel shows the relationship between token entropy under the pre-edit model and token-level editing difficulty. For each target token, we compute its Shannon entropy as a measure of model uncertainty before editing. We then quantify editing difficulty using a continuous local overlap metric between the edited output and the gold answer, where a larger value indicates that the corresponding local region is harder to rewrite correctly. The results show a clear increasing trend: higher-entropy tokens are consistently harder to edit, indicating that editability is highly non-uniform across long sequences. \textbf{Finding 1: Token entropy is positively correlated with editing difficulty, making it a reliable proxy for identifying hard-to-edit regions.}

The right panel further analyzes the semantic roles of tokens with different entropy levels. Function words, such as conjunctions, prepositions, pronouns, and auxiliary words, generally exhibit low entropy. In contrast, content-bearing tokens, including named entities, nouns, verbs, adjectives, and adverbs, show substantially higher entropy. This suggests that high-entropy regions tend to align with semantically informative positions rather than arbitrary difficult cases. \textbf{Finding 2: High-entropy tokens predominantly correspond to semantically informative content, while low-entropy tokens are mostly function words.}

Together, these findings expose a key limitation of existing long-sequence editing methods. By using fixed editing density and largely uniform optimization across the sequence, they spend unnecessary capacity on predictable, low-information regions while under-serving difficult and semantically important ones. This motivates DynEdit to allocate editing capacity according to token-level entropy, applying denser editing to high-entropy regions and focusing optimization on tokens that carry the core content of the target knowledge.

\section{Preliminaries}
\label{sec:preliminaries}

\textbf{Locate-and-Edit Formulation.} 
Given an input prompt $X=[x_1,\ldots,x_T]$ and a target sequence $Y=[y_1,\ldots,y_M]$, parameter-modifying knowledge editing aims to update the model $G_{\theta}$ such that it generates $Y$ while preserving unrelated behaviors \cite{meng2022locating}. Standard locate-and-edit methods achieve this by injecting a learnable memory shift $\delta \in \mathbb{R}^{d}$ into the hidden state $h_{p}^{l_e}$ at a chosen layer $l_e$ and position $p$. The optimal shift is typically found by minimizing the negative log-likelihood over the target sequence. However, relying on a single perturbation $\delta$ to explain a long, complex target sequence $Y$ introduces severe capacity bottlenecks.

\textbf{Sequential Window Editing.}
To address this limitation, sequence-level editors partition $Y$ into several chunks and optimize them sequentially \cite{2025arXiv250205628J}. At step $i$, a window $Y_{s_i:e_i}$ is optimized with prefix context $C_i = X \oplus Y_{<s_i}$ and edit position $p_i = |C_i|$.  Let $\{\delta_j^{*}\}_{j<i}$ denote previously optimized perturbations. The model execution at step $i$ is defined as:
\begin{equation}
    G_i(\delta_i)
    =
    G_{\theta}\!\left(
    \{h_{p_j}^{l_e} \leftarrow h_{p_j}^{l_e} + \delta_j^{*}\}_{j<i},
    h_{p_i}^{l_e} \leftarrow h_{p_i}^{l_e} + \delta_i
    \right),
\label{eq.7}
\end{equation}
where past edits are replayed and the current perturbation $\delta_i$ is injected at position $p_i$. The current perturbation is optimized via:
\begin{equation}
    \mathcal{L}_{\mathrm{win}}(\delta_i)
    =
    -\frac{1}{e_i-s_i+1}
    \sum_{t=s_i}^{e_i}
    \log P_{G_i(\delta_i)}(y_t \mid X,Y_{<t}),
\label{eq:window_loss}
\end{equation}
which uses a single perturbation $\delta_i$ to explain all tokens within the window.

\textbf{Problem Definition.} While the sequential paradigm improves long-form editing, existing methods rely on a \textit{fixed} window schedule (e.g., $\Delta_{\mathrm{fix}} = w-o$) and uniform token weighting (Eq.~\ref{eq:window_loss}). This formulation inherently ignores the heterogeneous semantics of language, leading to inefficient updates on simple regions and insufficient optimization for difficult, high-uncertainty tokens. Therefore, we study how to dynamically partition the target sequence and adaptively weight the optimization objective to improve the strength of long-form knowledge updates.

\section{Method}
\label{sec:method}

DynEdit is an entropy-guided framework for long-sequence knowledge editing. It keeps the standard locate-and-edit pipeline: each local edit first optimizes a hidden-state perturbation \(\delta_i\), and the resulting hidden targets are then projected back to model parameters by a backbone editor. The key difference is how these local edits are allocated. Instead of using a fixed window schedule and uniform token weights, DynEdit uses the predictive uncertainty of the pre-edit model to guide both window placement and token-level optimization.

\begin{figure}[h]
  \centering
  \includegraphics[width=\textwidth]{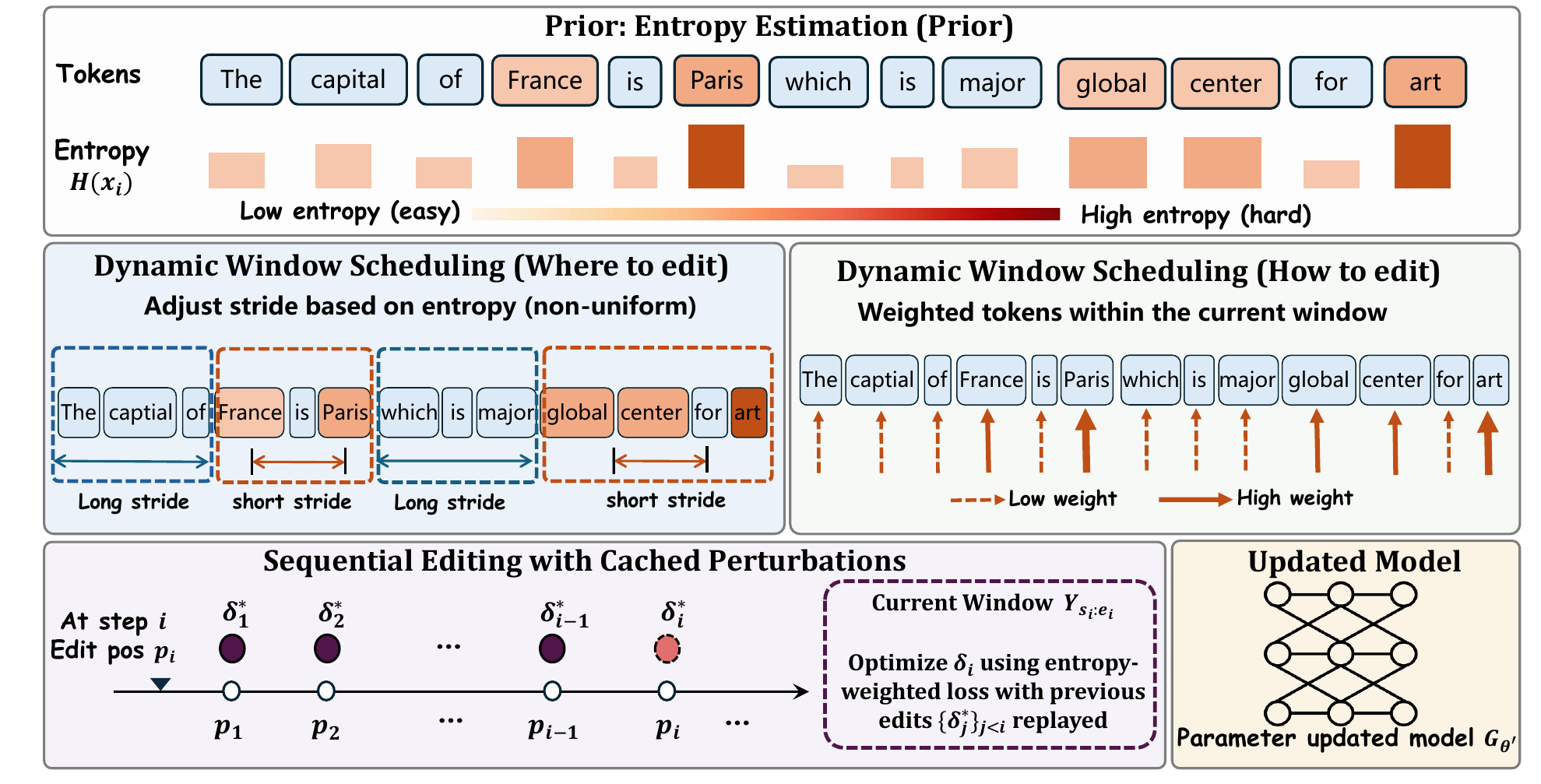}
  \caption{Overview of DynEdit. Our method consists of three components: token-level entropy estimation, entropy-guided dynamic window scheduling, and entropy-weighted sequential editing.
}\label{fig:fr}

  \vspace{-3mm}
\end{figure}

\subsection{Token-Level Entropy Estimation}
\label{subsec:entropy_estimation}

The first step of DynEdit is to identify which parts of the target sequence are more uncertain under the pre-edit model. This uncertainty map serves as the common signal for both dynamic scheduling and token-level weighting.

For each target position \(t\), DynEdit computes the pre-edit next-token
distribution under teacher forcing:
\begin{equation}
    p_t(v)=P_{G_{\theta}}(v\mid X,Y_{<t}), \qquad v\in\mathcal{V},
\label{eq.9}
\end{equation}
where \(\mathcal{V}\) is the vocabulary and \(Y_{<t}=[y_1,\ldots,y_{t-1}]\) is
the gold prefix. The token-level entropy is then defined as:
\begin{equation}
    H_t=-\sum_{v\in\mathcal{V}}p_t(v)\log p_t(v).
\label{eq.10}
\end{equation}

A larger \(H_t\) indicates that the pre-edit model is less certain about predicting \(y_t\), which, as shown in Sec.~\ref{sec:motivation}, often corresponds to tokens that are harder to edit and more semantically informative. All entropy values are computed once before editing and reused throughout the sequential editing process. This makes entropy estimation a lightweight pre-processing step rather than an additional optimization loop.

\subsection{Entropy-Guided Dynamic Scheduling}
\label{subsec:dynamic_scheduling}

Fixed-window editing uses a predefined stride throughout the target sequence, which implicitly assumes that all regions require the same editing density. DynEdit instead treats window scheduling as an adaptive allocation problem: the
stride for the next editing step is determined by the difficulty of the current local region.

At step \(i\), let \(s_i\) denote the start index of the current editing window.
DynEdit first constructs a candidate window
\begin{equation}
    \mathcal{W}_i = Y_{s_i:e_i}, \qquad
    e_i=\min(s_i+w-1,M),
\label{eq.11}
\end{equation}
where \(w\) is the base window size. This window is used to estimate the local editing difficulty around position \(s_i\).

The local difficulty score is defined as the average entropy within the window:
\begin{equation}
    \bar{H}_i=\frac{1}{e_i-s_i+1} \sum_{t=s_i}^{e_i}H_t.
\label{eq.12}
\end{equation}
To make the score comparable across different target sequences, we use the sequence-level mean entropy as an adaptive reference:
\begin{equation}
    \tau=\frac{1}{M}\sum_{t=1}^{M}H_t.
\label{eq.13}
\end{equation}
The relative difficulty of the current window is then
\begin{equation}
    d_i=\bar{H}_i-\tau.
\label{eq.14}
\end{equation}
A positive \(d_i\) indicates that the current region is more uncertain than the average region in the same target sequence, while a negative \(d_i\) indicates an easier region.

DynEdit maps this relative difficulty to the next stride through a monotone decreasing scheduling function:
\begin{equation}
    \Delta_i = \Phi(d_i),
    \qquad
    \frac{\partial \Phi(d)}{\partial d} \le 0.
\label{eq.15}
\end{equation}
Here, \(\Phi\) denotes an entropy-conditioned stride controller that maps the relative difficulty of the current window to the forward step size of the next window. Since \(\Phi\) is monotone decreasing, larger \(d_i\) produces a smaller \(\Delta_i\). Thus, harder regions induce smaller strides and receive denser window coverage. The next window start is updated by:
\begin{equation}
    s_{i+1}=s_i+\Delta_i.
\label{eq.16}
\end{equation}

This formulation generalizes fixed-stride editing. If \(\Phi(d)\) is a constant function, DynEdit reduces to the fixed-window schedule used by prior methods. When \(\Phi(d)\) decreases with \(d\), the editing trajectory becomes information-adaptive: high-entropy regions are revisited with greater local density, while low-entropy regions consume fewer editing steps.

\subsection{Entropy-Guided Token-Level Optimization}
\label{subsec:entropy_objective}

Dynamic scheduling determines \emph{where} to edit, but not \emph{how strongly} each token contributes to the update. Since tokens within a window differ in importance and difficulty \cite{fadeeva2024fact, liu2025token}, DynEdit uses entropy to modulate token-level optimization.

\textbf{Definition 1 (Gradient Dispersion).}
Given a sequence of tokens $\mathcal{T}$, let $\{g_i\}_{i \in \mathcal{T}}$ be their respective loss gradients. The gradient dispersion is defined as $\mathcal{D} = 1 - \frac{1}{|\mathcal{T}|(|\mathcal{T}|-1)} \sum_{i \neq j} \cos(g_i, g_j)$, measuring the degree of directional conflict. A higher $\mathcal{D}$ implies severe gradient interference during aggregation.

\textbf{Proposition 1 (Gradient Dilution in Shared Perturbation).} Let $\mathcal{T}$ be partitioned into a critical token set $\mathcal{C}$ and a background set $\mathcal{B}$, with their sum of gradients denoted as $g_c$ and $g_b$, respectively. Under uniform aggregation $g_{\text{uni}} \propto g_c + g_b$, if the background tokens exhibit high dispersion $\mathcal{D}_{\mathcal{B}}$, $g_b$ introduces a dominant orthogonal noise component, strictly upper-bounding the alignment $\cos(g_{\text{uni}}, g_c)$. Conversely, an importance-weighted aggregation $g_{\text{w}} \propto \alpha g_c + \beta g_b$ (with $\alpha > \beta > 0$) monotonically increases the lower bound of $\cos(g_{\text{w}}, g_c)$ compared to $g_{\text{uni}}$, thereby rescuing the critical update direction. The formal proof is provided in Appendix.

In long sequences, critical tokens are sparse and surrounded by many background tokens, leading to high dispersion. Under uniform aggregation, this biases updates toward majority directions and weakens critical gradients. DynEdit assigns entropy-based weights ($\alpha \gg \beta$) to emphasize high-uncertainty tokens, improving alignment with critical directions.

At step \(i\), DynEdit edits the last position of the prefix
\(C_i=X\oplus Y_{<s_i}\), namely \(p_i=|C_i|\). The edited execution \(G_i(\delta_i)\) is defined as in Eq.~\ref{eq.7}, with previous perturbations replayed.

\textbf{Token coefficients.}
DynEdit normalizes entropy within the target sequence:
\begin{equation}
    \hat{H}_t=
    \frac{H_t-H_{\min}}{H_{\max}-H_{\min}+\epsilon},
\label{eq.17}
\end{equation}
where \(H_{\min}\) and \(H_{\max}\) are the minimum and maximum token entropy values in the target sequence, and \(\epsilon\) is a small constant. The token coefficient is then:
\begin{equation}
\omega_t = \omega_0 + \kappa \hat{H}_t,
\label{eq.18}
\end{equation}
where \(\omega_0>0\) is the base coefficient and \(\kappa>0\) controls the strength of entropy modulation. Thus, high-entropy tokens receive larger coefficients while all tokens remain included in the objective.

\textbf{Focus and suffix coefficients.}
At step \(i\), DynEdit optimizes over the remaining suffix \(Y_{s_i:M}\), while using the current window \(\mathcal{W}_i=Y_{s_i:e_i}\) as the focus region. For
each \(t\ge s_i\), it assigns
\begin{equation}
    \lambda_{i,t}=
    \begin{cases}
        \omega_t, & s_i\le t\le e_i,\\
        \rho_{i,t}, & e_i<t\le M,
    \end{cases}
\label{eq.19}
\end{equation}
where \(\rho_{i,t}\) is a weak suffix coefficient. The focus weights strengthen optimization on difficult local tokens, while the suffix weights provide a future-aware constraint for later generation.

\textbf{Optimization objective.}
Using these coefficients, DynEdit optimizes
\begin{equation}
    \mathcal{L}_{\mathrm{Dyn}}(\delta_i)
    =
    -\frac{1}{Z_i}
    \sum_{t=s_i}^{M}
    \lambda_{i,t}
    \log P_{G_i(\delta_i)}(y_t\mid X,Y_{<t})
    +
    \gamma
    \frac{\|\delta_i\|_2}{\|h_{p_i}^{l_e}\|_2+\epsilon},
\label{eq.20}
\end{equation}
where \(Z_i=\sum_{t=s_i}^{M}\lambda_{i,t}\) normalizes the weighted loss. $\gamma$ controls the norm regularization, inheriting from the backbone editor. $\epsilon$ is a small constant for numerical stability. The optimized perturbation is
\begin{equation}
\delta_i^{*}=\arg\min_{\delta_i}\mathcal{L}_{\mathrm{Dyn}}(\delta_i).
\label{eq.21}
\end{equation}

This objective extends fixed-window editing in two ways: it evaluates the remaining suffix \(Y_{s_i:M}\) to preserve future continuation, and it assigns entropy-dependent coefficients inside the focus window to increase the influence of difficult, information-dense tokens. Thus, DynEdit improves both local token optimization and downstream sequence consistency.

\section{Experiments}
In this section, we perform a wide range of experiments to evaluate DynEdit, including editing performance, ablation study, general ability and length analysis.

\subsection{Editing performance}

In this section, we conduct editing experiments on two based LLMs using three datasets. Based on Table \ref{tab:1}, we can obtain the following observation.

\begin{table}[]
\caption{Editing performance (\%) of DynEdit against baselines on three benchmarks. AnyEdit and DynEdit denote the variants built on top of MEMIT, while AnyEdit* and DynEdit* denote the variants built on top of UnKE. }
\setlength{\tabcolsep}{1.7pt}
\scriptsize
\begin{tabular}{cc|cccccc|cccccc|ccc}
\hline
\multicolumn{2}{c|}{Datasets} & \multicolumn{6}{c|}{UnKE} & \multicolumn{6}{c|}{CounterFact} & \multicolumn{3}{c}{MQuAKE} \\ \hline
\multicolumn{1}{c|}{} &  & \multicolumn{3}{c|}{Original} & \multicolumn{3}{c|}{Paraphrase} & \multicolumn{3}{c|}{Original} & \multicolumn{3}{c|}{Paraphrase} & \multicolumn{3}{c}{Original} \\ \cline{3-17} 
\multicolumn{1}{c|}{\multirow{-2}{*}{Base Model}} & \multirow{-2}{*}{Methods} & BLEU & BERT & \multicolumn{1}{c|}{R-L} & BLEU & BERT & R-L & BLEU & BERT & \multicolumn{1}{c|}{R-L} & BLEU & BERT & R-L & BLEU & BERT & R-L \\ \hline
\multicolumn{1}{c|}{} & Pre-edited & 20.09 & 71.38 & \multicolumn{1}{c|}{25.49} & 19.8 & 71.29 & 25.31 & 13.89 & 68.33 & \multicolumn{1}{c|}{18.76} & 15.81 & 42.22 & 13.18 & 18.84 & 69.79 & 21.16 \\
\multicolumn{1}{c|}{} & FT-L \cite{hu2021lora} & 9.83 & 40.31 & \multicolumn{1}{c|}{11.39} & 3.24 & 37.29 & 8.51 & 6.53 & 42.89 & \multicolumn{1}{c|}{13.12} & 4.34 & 31.44 & 5.24 & 7.71 & 45.87 & 12.99 \\
\multicolumn{1}{c|}{} & MEND \cite{mitchellfast} & 21.08 & 68.73 & \multicolumn{1}{c|}{29.24} & 16.63 & 64.11 & 28.05 & 14.83 & 68.81 & \multicolumn{1}{c|}{30.30} & 10.63 & 41.56 & 10.95 & 13.37 & 67.85 & 22.48 \\
\multicolumn{1}{c|}{} & ROME \cite{meng2022locating} & 23.62 & 72.16 & \multicolumn{1}{c|}{23.74} & 18.8 & 70.54 & 22.39 & 13.24 & 72.90 & \multicolumn{1}{c|}{25.86} & 17.65 & 43.59 & 12.37 & 26.03 & 70.10 & 21.07 \\
\multicolumn{1}{c|}{} & MEMIT \cite{meng2022mass} & 24.97 & 76.21 & \multicolumn{1}{c|}{30.03} & 22.51 & 74.5 & 28.29 & 31.88 & 75.83 & \multicolumn{1}{c|}{31.23} & 17.52 & 47.27 & 15.88 & 26.61 & 69.93 & 25.75 \\
\multicolumn{1}{c|}{} & AlphaEdit \cite{fang2024alphaedit} & 21.43 & 73.76 & \multicolumn{1}{c|}{26.65} & 20.38 & 72.85 & 25.88 & 23.52 & 72.54 & \multicolumn{1}{c|}{24.98} & 16.39 & 45.13 & 14.07 & 22.57 & 69.78 & 22.74 \\
\multicolumn{1}{c|}{} & AnyEdit \cite{2025arXiv250205628J} & 77.91 & 95.69 & \multicolumn{1}{c|}{90.36} & 66.29 & 92.38 & \textbf{80.51} & 86.34 & 97.86 & \multicolumn{1}{c|}{95.51} & 39.48 & 65.31 & 46.26 & 87.28 & 97.38 & \textbf{94.28} \\
\multicolumn{1}{c|}{} & \cellcolor[HTML]{ECF4FF}\textbf{DynEdit} & \cellcolor[HTML]{ECF4FF}\textbf{91.6} & \cellcolor[HTML]{ECF4FF}\textbf{98.31} & \multicolumn{1}{c|}{\cellcolor[HTML]{ECF4FF}\textbf{92.62}} & \cellcolor[HTML]{ECF4FF}\textbf{80.3} & \cellcolor[HTML]{ECF4FF}\textbf{95.56} & \cellcolor[HTML]{ECF4FF}78.56 & \cellcolor[HTML]{ECF4FF}\textbf{95.08} & \cellcolor[HTML]{ECF4FF}\textbf{98.82} & \multicolumn{1}{c|}{\cellcolor[HTML]{ECF4FF}\textbf{96.11}} & \cellcolor[HTML]{ECF4FF}\textbf{44.88} & \cellcolor[HTML]{ECF4FF}\textbf{66.52} & \cellcolor[HTML]{ECF4FF}\textbf{48.02} & \cellcolor[HTML]{ECF4FF}\textbf{92.74} & \cellcolor[HTML]{ECF4FF}\textbf{98.26} & \cellcolor[HTML]{ECF4FF}93.11 \\
\multicolumn{1}{c|}{} & UnKE \cite{deng2024everything} & 93.2 & 98.09 & \multicolumn{1}{c|}{92.94} & 78.77 & 93.33 & 78.94 & 98.26 & 99.61 & \multicolumn{1}{c|}{98.25} & 38.63 & 60.28 & 34.4 & 95.04 & 98.93 & \textbf{94.4} \\
\multicolumn{1}{c|}{} & AnyEdit* & \textbf{99.54} & \textbf{99.79} & \multicolumn{1}{c|}{99.62} & 72.96 & 91.63 & 77.48 & \textbf{99.93} & \textbf{99.99} & \multicolumn{1}{c|}{\textbf{99.96}} & 40.25 & 62.95 & 42.71 & 99.79 & 99.97 & 99.87 \\
\multicolumn{1}{c|}{\multirow{-11}{*}{Llama-3-8B}} & \cellcolor[HTML]{ECF4FF}\textbf{DynEdit*} & \cellcolor[HTML]{ECF4FF}99.19 & \cellcolor[HTML]{ECF4FF} 99.37 & \multicolumn{1}{c|}{\cellcolor[HTML]{ECF4FF}\textbf{99.8}} & \cellcolor[HTML]{ECF4FF}\textbf{79.94} & \cellcolor[HTML]{ECF4FF}\textbf{96.71} & \cellcolor[HTML]{ECF4FF}\textbf{81.57} & \cellcolor[HTML]{ECF4FF}98.26 & \cellcolor[HTML]{ECF4FF} 99.74 & \multicolumn{1}{c|}{\cellcolor[HTML]{ECF4FF} 99.04} & \cellcolor[HTML]{ECF4FF}\textbf{46.19} & \cellcolor[HTML]{ECF4FF}\textbf{63.93} & \cellcolor[HTML]{ECF4FF}\textbf{42.79} & \cellcolor[HTML]{ECF4FF}\textbf{99.98} & \cellcolor[HTML]{ECF4FF}\textbf{99.99} & \cellcolor[HTML]{ECF4FF}\textbf{99.96} \\ \hline
\multicolumn{1}{c|}{} & Pre-edited & 20.92 & 72.58 & \multicolumn{1}{c|}{27.64} & 20.51 & 72.18 & 26.6 & 28.15 & 69.66 & \multicolumn{1}{c|}{20.98} & 29.54 & 52.15 & 19.31 & 24.12 & 69.05 & 21.03 \\
\multicolumn{1}{c|}{} & FT-L & 12.41 & 44.02 & \multicolumn{1}{c|}{13.78} & 6.33 & 40.33 & 12.93 & 16.47 & 46.66 & \multicolumn{1}{c|}{14.63} & 10.18 & 32.34 & 12.31 & 13.20 & 47.47 & 15.75 \\
\multicolumn{1}{c|}{} & MEND & 30.37 & 69.49 & \multicolumn{1}{c|}{27.77} & 15.53 & 62.01 & 27.92 & 35.47 & 69.54 & \multicolumn{1}{c|}{25.47} & 30.03 & 52.86 & 22.81 & 25.37 & 69.40 & 32.39 \\
\multicolumn{1}{c|}{} & ROME & 43.32 & 74.73 & \multicolumn{1}{c|}{31.52} & 37.40 & 71.90 & 28.12 & 44.14 & 75.89 & \multicolumn{1}{c|}{36.42} & 31.34 & 55.67 & 25.79 & 42.26 & 72.18 & 35.61 \\
\multicolumn{1}{c|}{} & MEMIT & 45.18 & 78.1 & \multicolumn{1}{c|}{38.21} & 40.89 & 76.63 & 34.19 & 45.08 & 77.16 & \multicolumn{1}{c|}{38.95} & 32.99 & 56.15 & 25.8 & 41.54 & 71.61 & 35.46 \\
\multicolumn{1}{c|}{} & AlphaEdit & 49.75 & 80.57 & \multicolumn{1}{c|}{42.93} & 45.37 & 78.41 & 38.42 & 49.88 & 80.42 & \multicolumn{1}{c|}{45.35} & 34.65 & 56.94 & 27.74 & 45.14 & 72.76 & 39.83 \\
\multicolumn{1}{c|}{} & AnyEdit & 86.86 & 96.85 & \multicolumn{1}{c|}{91.68} & 64.94 & 90.11 & 71.47 & 88.52 & 97.29 & \multicolumn{1}{c|}{91.94} & 38.1 & 62.72 & 39.43 & 86.69 & 96.48 & 90.57 \\
\multicolumn{1}{c|}{} & \cellcolor[HTML]{ECF4FF}\textbf{DynEdit} & \cellcolor[HTML]{ECF4FF}\textbf{95.96} & \cellcolor[HTML]{ECF4FF}\textbf{99.71} & \multicolumn{1}{c|}{\cellcolor[HTML]{ECF4FF}\textbf{96.77}} & \cellcolor[HTML]{ECF4FF}\textbf{77.49} & \cellcolor[HTML]{ECF4FF}\textbf{93.27} & \cellcolor[HTML]{ECF4FF}\textbf{78.56} & \cellcolor[HTML]{ECF4FF}\textbf{95.66} & \cellcolor[HTML]{ECF4FF}\textbf{98.93} & \multicolumn{1}{c|}{\cellcolor[HTML]{ECF4FF}\textbf{96.08}} & \cellcolor[HTML]{ECF4FF}\textbf{45.61} & \cellcolor[HTML]{ECF4FF}\textbf{66.05} & \cellcolor[HTML]{ECF4FF}\textbf{43.09} & \cellcolor[HTML]{ECF4FF}\textbf{95.02} & \cellcolor[HTML]{ECF4FF}\textbf{98.82} & \cellcolor[HTML]{ECF4FF}\textbf{95.12} \\
\multicolumn{1}{c|}{} & UnKE & 91.72 & 96.85 & \multicolumn{1}{c|}{90.79} & 56.79 & 83.85 & 51.97 & 91.87 & 97.54 & \multicolumn{1}{c|}{91.11} & 38.45 & 59.15 & 29.16 & 88.04 & 95.15 & 86.31 \\
\multicolumn{1}{c|}{} & AnyEdit* & 96.13 & 98.85 & \multicolumn{1}{c|}{97.30} & 47.17 & 80.75 & 50.96 & 96.70 & 99.10 & \multicolumn{1}{c|}{97.48} & 32.02 & 57.29 & 31.94 & 97.44 & 99.49 & 97.70 \\
\multicolumn{1}{c|}{\multirow{-11}{*}{Qwen-2.5-7B}} & \cellcolor[HTML]{ECF4FF}\textbf{DynEdit*} & \cellcolor[HTML]{ECF4FF}\textbf{99.54} & \cellcolor[HTML]{ECF4FF}\textbf{99.71} & \multicolumn{1}{c|}{\cellcolor[HTML]{ECF4FF}\textbf{99.44}} & \cellcolor[HTML]{ECF4FF}\textbf{66.13} & \cellcolor[HTML]{ECF4FF}\textbf{88.24} & \cellcolor[HTML]{ECF4FF}\textbf{66.12} & \cellcolor[HTML]{ECF4FF}\textbf{99.08} & \cellcolor[HTML]{ECF4FF}\textbf{99.64} & \multicolumn{1}{c|}{\cellcolor[HTML]{ECF4FF}\textbf{99.09}} & \cellcolor[HTML]{ECF4FF}\textbf{39.82} & \cellcolor[HTML]{ECF4FF}\textbf{62.32} & \cellcolor[HTML]{ECF4FF}\textbf{36.87} & \cellcolor[HTML]{ECF4FF}\textbf{99.83} & \cellcolor[HTML]{ECF4FF}\textbf{99.9} & \cellcolor[HTML]{ECF4FF}\textbf{99.63} \\ \hline
\end{tabular}
\label{tab:1}
\vspace{-4mm}
\end{table}

\textbf{DynEdit consistently improves long-sequence editing performance across different backbones.}
Under the MEMIT backbone, DynEdit outperforms AnyEdit on both Llama-3-8B and Qwen-2.5-7B across nearly all datasets, achieving up to +12.5 BLEU on UnKE-Original and consistent gains on CounterFact and MQuAKE. These improvements persist under the stronger UnKE backbone, where DynEdit$^\ast$ further improves over AnyEdit$^\ast$ by up to +3.5 BLEU. This demonstrates that DynEdit provides stable gains independent of the underlying editor.

\textbf{DynEdit substantially improves robustness on paraphrased queries, indicating better knowledge generalization.}
On paraphrase evaluation, DynEdit achieves consistent gains over AnyEdit, with improvements of up to +14.0 BLEU on UnKE and +7.5 BLEU on CounterFact across both models. This suggests that DynEdit goes beyond surface-form fitting and better transfers edited knowledge to semantically equivalent inputs.

\textbf{DynEdit is particularly effective on challenging long-form and compositional editing tasks.}
The advantage becomes more pronounced on difficult settings such as MQuAKE and paraphrased long-form editing, where DynEdit improves performance by up to +8.3 BLEU. These results align with our design: by allocating more editing capacity to high-entropy regions, DynEdit better captures critical information and maintains sequence-level consistency. The full results on more editors and models (e.g., Qwen-3) are provided in the Appendix \ref{fll_results}.

\subsection{Diverse-Formatted Knowledge Editing Evaluation}
We further evaluate DynEdit on diverse-formatted knowledge using EditEverything, which covers multiple domains with varying structures, including mathematics, programming, poetry, news, and chemistry. Results are reported in Table~\ref{tab:2}.

\begin{table}[htbp]
\caption{Performance comparison (\%) between the DynEdit approach and baseline methods on long-form diverse-formatted knowledge. Knowledge types without underlining represent Rouge-L Score metrics,
while those with underlining indicate Bert Score metrics.}
\scriptsize
\centering
\begin{tabular}{l|ccccccccccc}
\hline
\multicolumn{1}{c|}{\textbf{Base Model}} & \textbf{Method} & {\ul \textbf{math}} & {\ul \textbf{code}} & {\ul \textbf{poetry}} & {\ul \textbf{news}} & {\ul \textbf{chemistry}} & \textbf{math} & \textbf{code} & \textbf{poetry} & \textbf{news} & \textbf{chemistry} \\ \hline
 & RoME & 82.34 & 72.63 & 65.45 & 69.08 & 80.86 & 66.7 & 58.17 & 21.83 & 28.87 & 69.97 \\
 & MEMIT & 96.67 & 84.3 & 76.78 & 83.63 & 93.6 & 77.36 & 60.03 & 20.8 & 33.9 & 72.29 \\
 & AlphaEdit & 96.41 & 83.28 & 76.13 & 80.1 & 93.16 & 77.36 & 58.7 & 24.31 & 31.52 & 65.11 \\
 & AnyEdit & 98.22 & 94.08 & 91.93 & 95.6 & 99.07 & 91.72 & 92.63 & 73.73 & 88.31 & 92.53 \\
\multirow{-5}{*}{Qwen2.5-7B} & \cellcolor[HTML]{ECF4FF}\textbf{DynEdit} & \cellcolor[HTML]{ECF4FF}\textbf{99.73} & \cellcolor[HTML]{ECF4FF}\textbf{99.11} & \cellcolor[HTML]{ECF4FF}\textbf{98.13} & \cellcolor[HTML]{ECF4FF}\textbf{98.98} & \cellcolor[HTML]{ECF4FF}\textbf{99.58} & \cellcolor[HTML]{ECF4FF}\textbf{99.07} & \cellcolor[HTML]{ECF4FF}\textbf{98.78} & \cellcolor[HTML]{ECF4FF}\textbf{96.39} & \cellcolor[HTML]{ECF4FF}\textbf{97.37} & \cellcolor[HTML]{ECF4FF}\textbf{99.23} \\ \hline
 & RoME & 79.82 & 70.47 & 58.83 & 64.49 & 70.6 & 35.23 & 39.13 & 24.5 & 22.31 & 31.77 \\
 & MEMIT & 83.01 & 77.73 & 67.71 & 70.03 & 88.62 & 41.12 & 35.41 & 21.3 & 29.67 & 46.2 \\
 & AlphaEdit & 82.38 & 76.85 & 65.57 & 68.78 & 88.58 & 42.63 & 37.68 & 24.43 & 33.5 & 48.78 \\
 & AnyEdit & 91.29 & 96.37 & 85.38 & 92.73 & 97.78 & 70.73 & 84.38 & 69.7 & 81.23 & 79.9 \\
\multirow{-5}{*}{Llama-3-8B} & \cellcolor[HTML]{ECF4FF}\textbf{DynEdit} & \cellcolor[HTML]{ECF4FF}\textbf{99.31} & \cellcolor[HTML]{ECF4FF}\textbf{98.93} & \cellcolor[HTML]{ECF4FF}\textbf{97.5} & \cellcolor[HTML]{ECF4FF}\textbf{97.81} & \cellcolor[HTML]{ECF4FF}\textbf{99.83} & \cellcolor[HTML]{ECF4FF}\textbf{97.76} & \cellcolor[HTML]{ECF4FF}\textbf{96.07} & \cellcolor[HTML]{ECF4FF}\textbf{95.49} & \cellcolor[HTML]{ECF4FF}\textbf{98.78} & \cellcolor[HTML]{ECF4FF}\textbf{99.13} \\ \hline
\end{tabular}
\label{tab:2}
\end{table}

\textbf{DynEdit consistently outperforms strong baselines across all domains and base models.} DynEdit achieves uniform improvements over AnyEdit on all five knowledge types under both Qwen2.5-7B and Llama-3-8B, with gains of up to +27.0 Rouge-L and consistent improvements in BERTScore. This demonstrates that DynEdit generalizes well across heterogeneous output formats and evaluation metrics.

\textbf{DynEdit shows larger gains on structurally complex and semantically dense knowledge.} The improvements are most pronounced on domains such as math, code, and poetry, where correct editing depends on a small number of critical tokens and strong sequence-level coherence. This aligns with our design: entropy-guided scheduling and weighting focus optimization on informative regions, leading to more accurate and consistent edits.

\subsection{Ablation Study}

\begin{figure*}[h]
    \centering
    \begin{minipage}[t]{0.48\textwidth}
        \vspace{0pt}  
        \centering
        \includegraphics[width=\linewidth]{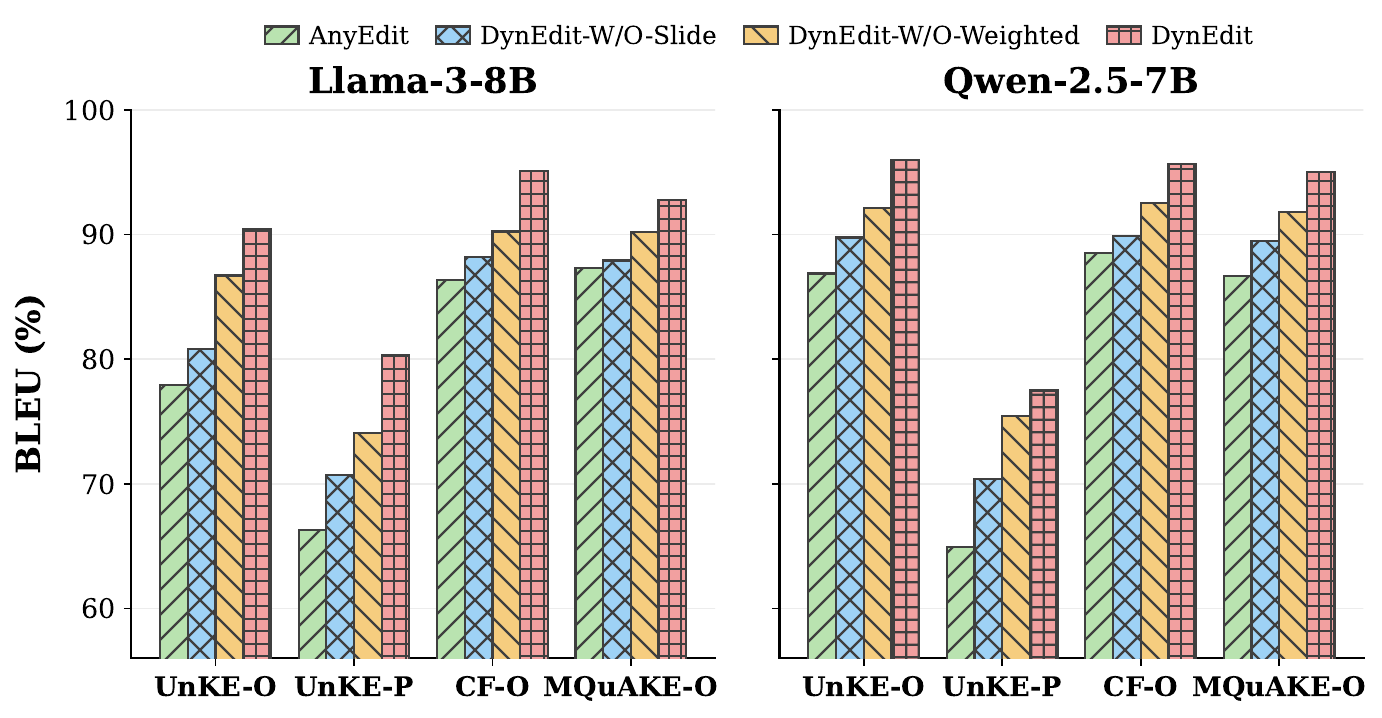}
        \caption{Ablation study of DynEdit. -O denotes Original BLEU score; -P denotes Paraphrase BLEU score.}
        \label{fig:4}
    \end{minipage}
    \hfill
    \begin{minipage}[t]{0.48\textwidth}
        \vspace{0pt}  
        \centering
        \captionof{table}{Performance comparison (\%) between DynEdit and baselines on general benchmarks.}
        \label{tab:3}
        \scriptsize
        \setlength{\tabcolsep}{3.8pt}
        \begin{tabular}{|c|c|c|c|c|}
        \hline
        \textbf{LLM} & \textbf{Method} & \textbf{MMLU} & \textbf{IFEval (S)} & \textbf{IFEval (L)} \\ \hline
         & Pre-edited & 67.16$\pm$0.38 & 69.50$\pm$1.98 & 75.60$\pm$1.85 \\ \cline{2-5} 
         & AnyEdit & 65.81$\pm$0.12 & 64.21$\pm$0.12 & 64.21$\pm$0.65 \\ \cline{2-5} 
         & \cellcolor[HTML]{ECF4FF}DynEdit & \cellcolor[HTML]{ECF4FF}\textbf{66.03$\pm$0.12} & \cellcolor[HTML]{ECF4FF}63.01$\pm$0.37 & \cellcolor[HTML]{ECF4FF}\textbf{71.56$\pm$0.61} \\ \cline{2-5} 
         & AnyEdit* & 65.72$\pm$0.12 & \textbf{70.13$\pm$0.62} & 75.91$\pm$0.58 \\ \cline{2-5} 
        \multirow{-5}{*}{Llama3} & \cellcolor[HTML]{ECF4FF}DynEdit* & \cellcolor[HTML]{ECF4FF}65.78$\pm$0.12 & \cellcolor[HTML]{ECF4FF}69.95$\pm$0.43 & \cellcolor[HTML]{ECF4FF}\textbf{75.99$\pm$0.59} \\ \hline
         & Pre-edited & 73.45$\pm$0.36 & 70.24$\pm$1.97 & 73.57$\pm$1.90 \\ \cline{2-5} 
         & AnyEdit & 72.21$\pm$0.11 & 65.56$\pm$0.65 & 68.74$\pm$0.63 \\ \cline{2-5} 
         & \cellcolor[HTML]{ECF4FF}DynEdit & \cellcolor[HTML]{ECF4FF}\textbf{72.33$\pm$0.11} & \cellcolor[HTML]{ECF4FF}\textbf{66.33$\pm$0.63} & \cellcolor[HTML]{ECF4FF}\textbf{70.26$\pm$0.60} \\ \cline{2-5} 
         & AnyEdit* & 73.28$\pm$0.11 & 70.89$\pm$0.62 & 73.71$\pm$0.60 \\ \cline{2-5} 
        \multirow{-5}{*}{Qwen2.5} & \cellcolor[HTML]{ECF4FF}DynEdit* & \cellcolor[HTML]{ECF4FF}\textbf{73.53$\pm$0.11} & \cellcolor[HTML]{ECF4FF}\textbf{70.93$\pm$0.73} & \cellcolor[HTML]{ECF4FF}\textbf{74.55$\pm$0.60} \\ \hline
        \end{tabular}
    \end{minipage}
\vspace{-3mm}
\end{figure*}

We evaluate the contribution of each component by removing the dynamic sliding strategy and the entropy-guided weighting module. Results on UnKE, CounterFact, and MQuAKE are shown in Figure~\ref{fig:4}.

\textbf{Both dynamic sliding and entropy-guided weighting are important for DynEdit.} Removing either component consistently reduces performance across datasets and models, confirming that the two modules jointly contribute to long-sequence editing. Nevertheless, the ablated variants still outperform AnyEdit in most settings, suggesting that even partial entropy-aware allocation is more effective than fixed uniform editing.

\textbf{Removing the sliding strategy causes a larger performance drop, suggesting that entropy-guided weighting depends on adaptive local editing.} Without sliding, each chunk mainly observes the suffix from a fixed local position, which substantially weakens the effect of entropy-guided token weighting. As a result, the weighting mechanism cannot fully focus optimization on the most informative regions. This explains why removing sliding hurts the full system more severely.

\subsection{Analysis of General Ability}

To evaluate whether long-sequence editing harms the model's general ability, we further test the edited models on MMLU \cite{hendrycks2020measuring} and IFEval \cite{zhou2023instruction}. We compare both the MEMIT-based setting (AnyEdit vs.\ DynEdit) and the UnKE-based setting (AnyEdit$^\ast$ vs.\ DynEdit$^\ast$), with results shown in Table~\ref{tab:3}.

\textbf{DynEdit preserves general ability while maintaining competitive or improved performance over baselines.} Across both base models and editing backbones, DynEdit matches or slightly outperforms AnyEdit on most metrics. Under the MEMIT setting, DynEdit consistently improves MMLU and IFEval over AnyEdit, while under the stronger UnKE backbone, DynEdit$^\ast$ remains competitive and achieves comparable or better results. These results indicate that entropy-guided editing does not introduce additional degradation and maintains the model’s general capabilities.


\subsection{Length-wise Robustness Analysis}
\label{subsec:length_robustness}

We evaluate robustness w.r.t.\ target length on UnKE by grouping samples into four quantiles (Q1–Q4) from short to long. We compare AnyEdit and DynEdit within each bucket under both original and paraphrased prompts, and additionally visualize position-wise editing error along the target sequence.

\begin{figure}[h]
  \centering
  \includegraphics[width=\textwidth]{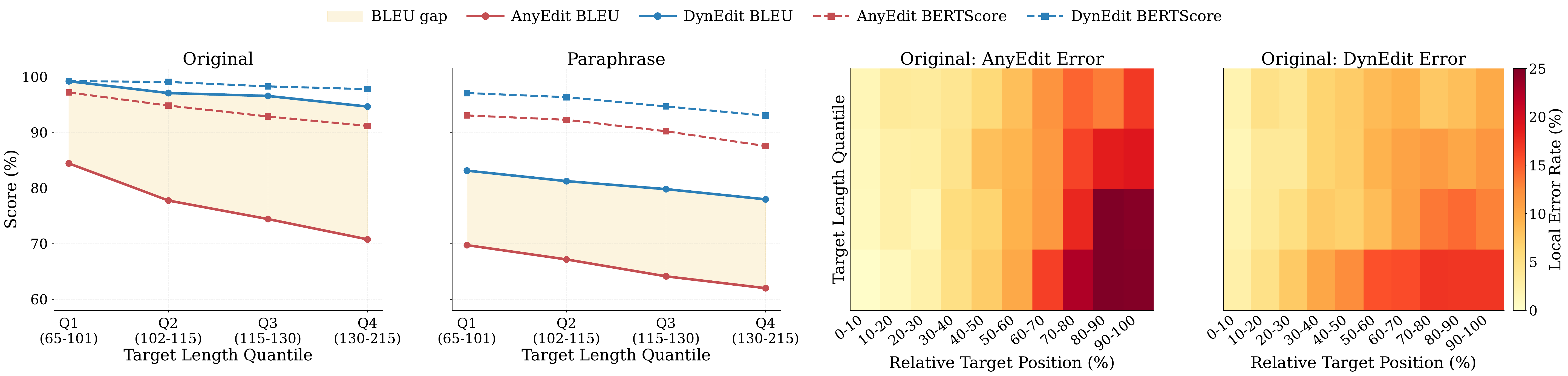}
  \caption{Length-wise robustness and position-wise error analysis.}
  \label{fig:5}
\vspace{-4mm}
\end{figure}

\textbf{DynEdit achieves stronger robustness on long targets by reducing error accumulation.}
Performance degrades with increasing length for both methods, but DynEdit shows a significantly flatter drop across Q1–Q4. Position-wise analysis further reveals that AnyEdit accumulates errors toward later tokens, especially for long sequences, while DynEdit maintains lower and more stable error profiles. This suggests that entropy-guided dynamic allocation better mitigates long-range error propagation by focusing optimization on difficult regions. 

The appendix provides more experimental results, including full experimental results \ref{fll_results}, parameter analysis \ref{hyper}, editing efficiency analysis \ref{efficiency}, batch editing \ref{batch}, and case studies \ref{case}.

\section{Conclusion}
\label{sec:conclusion}

We study long-sequence knowledge editing, where existing methods allocate editing effort uniformly despite heterogeneous token difficulty. To address this, we propose DynEdit, an entropy-guided framework that dynamically adjusts both editing windows and token-level optimization weights. Experiments across multiple benchmarks and models show that DynEdit consistently outperforms strong baselines, particularly on long and paraphrased targets, while preserving general model behavior. Overall, DynEdit provides a simple and effective approach to adapt editing allocation to the difficulty structure of long sequences.

\section{Acknowledgement}
This research is supported by the National Natural Science Foundation of China (62072052).

\bibliographystyle{plainnat} 
\bibliography{main}

\newpage

\appendix

\section{Related Works}
Large language models (LLMs) have demonstrated remarkable capabilities across diverse tasks \cite{wolf2019huggingface,radford2019language}, yet they inevitably store outdated, incorrect, or conflicting factual knowledge within their parameters \cite{zhang2024comprehensive}. Model editing \cite{meng2022locating} has emerged as a promising approach to address this challenge , enabling precise updates to model behavior toward target facts without the computational overhead of full retraining or fine-tuning \cite{hu2021lora}.

Existing knowledge editing methods can be broadly grouped into two paradigms. Context-augmentation methods inject edited knowledge through prompts or external contexts, making them suitable for black-box settings. IKE \cite{zheng2023can} and ICE \cite{qi2024context} leverage in-context learning for knowledge editing, while EditCoT \cite{EditCoT} and CoT2Edit \cite{fu2026learning} extend this paradigm by incorporating chain-of-thought reasoning. SERAC \cite{mitchell2022memory} retrieves updates from external memory, and GRACE \cite{hartvigsen2024aging} employs discrete key-value adaptors for lifelong editing. However, these methods face challenges in efficiently editing large amounts of knowledge due to limitations in context window length and parameter count.

Parameter-modifying methods directly update model parameters by locating and modifying knowledge-relevant representations. Meta-learning-based approaches such as MEND \cite{mitchellfast} train hypernetworks to predict parameter updates, while ROME \cite{meng2022locating} introduces causal tracing \cite{pearl2001direct, vig2020investigating} to identify knowledge-relevant layers and employs rank-one model editing to modify feedforward layer weights. MEMIT \cite{meng2022mass} extends this to batch editing, enabling simultaneous updates of multiple factual associations. AlphaEdit \cite{fang2024alphaedit} further addresses lifelong editing scenarios by projecting updates onto the null space of previously edited knowledge, mitigating catastrophic forgetting. Besides, other works \cite{zhang2024locate, zhang2024knowledge, li2024pmet, ma2024perturbation, gu2024model} have also extensively explored methods for parameter modification. 

Despite substantial progress, most parameter‑modifying methods are designed for localized editing, computing updates from a small number of fact‑relevant positions. While effective for short triplet‑style facts, this design is inadequate for long‑sequence knowledge editing (e.g., mathematical derivations, code snippets), where a single editing position cannot cover the dependencies distributed over the entire answer.

To address this limitation, recent methods have extended editing from a single position to multiple target positions. UnKE \cite{deng2024everything} discards the explicit knowledge locating step and treats the first few transformer layers as the key generator, optimizing over long target sequences by updating all parameters within a single layer. AnyEdit \cite{2025arXiv250205628J} introduces an autoregressive sliding-window paradigm that decomposes long-form knowledge into sequential chunks and iteratively edits the key token in each chunk, establishing a theoretical foundation based on the chain rule of mutual information. 

However, such sequence-level editors still rely on difficulty-agnostic allocation and suffer from token-level optimization imbalance, overlooking the non-uniform nature of long-form knowledge \cite{li2026adacurl, dai2026adaptive}. These deficiencies directly motivate our proposed DynEdit framework.

\section{Proof of Proposition 1}
\label{Proof}

\textbf{Proof (Sketch).}
Let $g_c = \sum_{i \in \mathcal{C}} g_i$ and $g_b = \sum_{j \in \mathcal{B}} g_j$. The uniformly aggregated gradient is
\[
g_{\text{uni}} = g_c + g_b.
\]

We analyze the alignment between $g_{\text{uni}}$ and $g_c$:
\[
\cos(g_{\text{uni}}, g_c) = \frac{\langle g_c + g_b, g_c \rangle}{\|g_c + g_b\| \cdot \|g_c\|}.
\]

Expanding the numerator:
\[
\langle g_c + g_b, g_c \rangle = \|g_c\|^2 + \langle g_b, g_c \rangle.
\]

When the background gradients exhibit high dispersion, their pairwise cosine similarities are low, implying that $g_b$ contains substantial components orthogonal to $g_c$. We decompose:
\[
g_b = g_b^{\parallel} + g_b^{\perp},
\]
where $g_b^{\parallel}$ is parallel to $g_c$ and $g_b^{\perp}$ is orthogonal.

Thus:
\[
\langle g_b, g_c \rangle = \langle g_b^{\parallel}, g_c \rangle,
\]
and $\|g_c + g_b\|^2 = \|g_c + g_b^{\parallel}\|^2 + \|g_b^{\perp}\|^2$.

As dispersion $\mathcal{D}_{\mathcal{B}}$ increases, $\|g_b^{\perp}\|$ grows, which increases the denominator while not contributing to the numerator. Therefore, $\cos(g_{\text{uni}}, g_c)$ decreases, establishing an upper bound controlled by the magnitude of orthogonal noise.

Now consider the weighted aggregation:
\[
g_{\text{w}} = \alpha g_c + \beta g_b, \quad \alpha > \beta > 0.
\]

Similarly,
\[
\cos(g_{\text{w}}, g_c) = \frac{\alpha \|g_c\|^2 + \beta \langle g_b, g_c \rangle}{\|\alpha g_c + \beta g_b\| \cdot \|g_c\|}.
\]

Compared to the uniform case, increasing $\alpha$ amplifies the contribution of $\|g_c\|^2$ in the numerator, while scaling down $\beta$ reduces the impact of $\|g_b^{\perp}\|$ in the denominator. This improves the alignment with $g_c$.

More formally, for fixed $g_c, g_b$, $\cos(g_{\text{w}}, g_c)$ is monotonically increasing in $\alpha / \beta$, establishing a tighter lower bound compared to uniform aggregation.

\section{Experimental Supplements}
In this section, we provide further experimental details, including datasets, comparison methods, and evaluation metrics. Furthermore, we present additional experimental results, including efficiency analysis, results on AlphaEdit, and results on Qwen3-8B.

\subsection{Datasets and Baselines}
\label{dataset}
\subsubsection{Datasets}
We evaluate our method on four knowledge editing benchmarks covering both structured and unstructured, long-form knowledge:

\textbf{UnKEBench} \cite{deng2024everything}. A counterfactual unstructured knowledge editing dataset containing 1,000 long-form texts. Unlike triplet-based datasets, each instance pairs a natural language question with a multi-sentence answer, requiring the model to edit complex, information-dense knowledge while avoiding interference from pre-trained memory.

\textbf{CounterFact} \cite{meng2022locating}. A widely used structured editing benchmark where each edit is a counterfactual triplet (subject, relation, object). Following prior work (Meng et al., 2023; Jiang et al., 2025), we evaluate both original and paraphrased query performance.

\textbf{MQuAKE} \cite{zhong2023mquake}. A multi-hop knowledge editing dataset that assesses whether edited facts can correctly propagate through complex reasoning chains. Each edit involves a sequence of interrelated facts, testing the model’s ability to maintain consistency across multiple knowledge updates.

\textbf{EditEverything} \cite{2025arXiv250205628J}. A recently proposed benchmark for long-form, diverse-format knowledge editing. It spans multiple domains—mathematics, poetry, news, code, and chemistry—with target sequences up to 458 tokens, enabling evaluation of editing methods on realistic, format-rich content.

\subsubsection{Baselines}

We compare our method against a range of representative knowledge editing approaches:

\textbf{FT-L} \cite{hu2021lora}. Fine-tuning with LoRA applied to the last transformer layer (or a small subset of parameters). It serves as a simple but computationally efficient baseline for knowledge updating.

\textbf{MEND} \cite{mitchellfast}. A hypernetwork-based editor that learns to predict gradient updates for model parameters, enabling fast single-sample editing without full fine-tuning.

\textbf{MEMIT} \cite{meng2022mass}. A locate-then-edit method that performs batch editing by updating feedforward layers across multiple transformer blocks using a closed-form least-squares solution.

\textbf{AlphaEdit} \cite{fang2024alphaedit}. An extension of MEMIT that projects parameter updates onto the null space of previously edited knowledge, designed for lifelong editing scenarios.

\textbf{UnKE} \cite{deng2024everything}. An unstructured knowledge editing method that discards explicit token localization and optimizes over the entire target sequence by updating all parameters within a single transformer layer.

\textbf{AnyEdit} \cite{2025arXiv250205628J}. A sliding-window autoregressive editing paradigm that decomposes long-form knowledge into chunks and iteratively edits the last token of each chunk, enabling longer-sequence updates than single-position methods.

\subsection{Experimental setup}
\label{subsec:experimental_setup}

All experiments are conducted on a single NVIDIA H20 GPU (96G). We evaluate DynEdit on the same benchmarks and evaluation protocol as AnyEdit. Specifically, we use Llama3-8B-Instruct and Qwen-series models as base LLMs, and evaluate editing performance on original queries, paraphrased queries, and locality-related sub-questions when available. Following prior long-form editing evaluation, we report lexical matching metrics including BLEU and ROUGE, together with BERT Score \cite{devlin2019bert} computed by the \texttt{all-MiniLM-L6-v2} \cite{wang2020minilm, reimers2019sentence} sentence encoder. During generation, we use a maximum generation length of 512 tokens and a near-deterministic decoding temperature. Our codes are available at \url{https://github.com/FredJDean/DynEdit}.

\paragraph{Backbone editors.}
DynEdit is implemented on top of the same backbone editors used by AnyEdit. We denote the MEMIT-based instantiation as DynEdit, and the UnKE-based instantiation as DynEdit$^{*}$, following the notation of AnyEdit and AnyEdit$^{*}$. For fair comparison, DynEdit and DynEdit$^{*}$ keep the backbone-specific update rules unchanged. That is, DynEdit uses the same constrained regression update as the MEMIT-based AnyEdit setting, while DynEdit$^{*}$ follows the same layer-wise optimization procedure as the UnKE-based AnyEdit$^{*}$ setting. Our method only changes how target positions are scheduled and how token-level losses are weighted during hidden-target construction.

\paragraph{MEMIT-based DynEdit.}
For the MEMIT-based setting, we follow the AnyEdit configuration for the selected editing layers, update module, and optimization hyperparameters. Edits are applied to the MLP down-projection modules of layers 4--8. The fact token is set to the subject-last token for Llama-series models and the last token for Qwen-series models. We use 25 gradient steps to optimize the hidden target, with a value learning rate of $5\times 10^{-1}$, clamp norm factor of 4, weight decay of $10^{-3}$, and KL regularization factor of 0.0625. The covariance statistics are computed on Wikipedia with 100K samples in \texttt{float32}, and the MEMIT update weight is set to 15000, consistent with AnyEdit.

\paragraph{UnKE-based DynEdit$^{*}$.}
For the UnKE-based setting, DynEdit$^{*}$ follows the AnyEdit$^{*}$ implementation protocol. The optimization uses AdamW with learning rate $2\times 10^{-4}$, 50 optimization steps for the layer-wise update, 25 gradient steps for hidden-target optimization, value learning rate $5\times 10^{-1}$, clamp norm factor of 4, and weight decay of $10^{-3}$. We sample 20 auxiliary data points to constrain the update, following the AnyEdit$^{*}$ setting. The loss layer is set to layer 31 for Llama3-8B-Instruct, layer 27 for Qwen2.5-7B-Instruct, and layer 35 for Qwen3-8B.

\paragraph{Dynamic sliding configuration.}
The main difference between DynEdit/DynEdit$^{*}$ and AnyEdit/AnyEdit$^{*}$ lies in the sequential target scheduling. AnyEdit uses fixed chunks with no adaptive overlap, whereas DynEdit computes token-level entropy under the pre-edit model and adjusts the stride according to the entropy of the current window. For a current window, if its mean entropy is larger than the mean entropy of the whole target sequence, the window is regarded as a difficult region and DynEdit uses a denser stride. Otherwise, the method advances with the default stride determined by the configured window size and overlap. In our implementation, the default overlap is set to 0, matching AnyEdit, and dense regions are assigned a smaller stride. For Llama-series models, the dense stride is set according to the Llama-specific DynEdit configuration; for Qwen-series models, the dense stride is set to one third of the window size. This adaptive sliding strategy is applied consistently to both DynEdit and DynEdit$^{*}$.

\paragraph{Window and horizon.}
For the fixed-window AnyEdit baselines, we follow the original chunk configuration: 40 tokens for Llama-series models and 50 tokens for Qwen-series models. For DynEdit variants, the base window size is specified by the corresponding DynEdit hyperparameter file, while the stride is dynamically adjusted by the entropy criterion above. At each editing step, DynEdit constructs the prefix using all target tokens before the current start position and optimizes the remaining suffix as the future horizon. Tokens inside the current focus window receive entropy-guided weights, while future suffix tokens provide additional regularization signals. This keeps the backbone editor unchanged while replacing fixed chunking with entropy-guided sequential allocation.

\subsection{Results of DynEdit in more Editors and Qwen3}
\label{fll_results}
We provide additional results on UnKEBench to further evaluate the generality of DynEdit across different editors and base models, including AlphaEdit and Qwen3-8B (Table~\ref{all}).

\begin{table}[h]
\scriptsize
\caption{DynEdit against baselines in all metrics of UnKEBench, including additional AlphaEdit editor as well as Qwen3-8B base model. The left of "/" is Original and the right is Paraphrase.}
\begin{tabular}{ccccccc}
\hline
 & \multicolumn{4}{c}{\textbf{Lexical Similarity}} & \textbf{Semantic Similarity} & \textbf{Sub Questions} \\ \cline{2-7} 
\multirow{-2}{*}{\textbf{Method}} & \textbf{BLEU} & \textbf{ROUGE-1} & \textbf{ROUGE-2} & \textbf{ROUGE-L} & \textbf{BERT Score} & \textbf{ROUGE-L} \\ \hline
\multicolumn{7}{c}{\textbf{Based on Llama3-8B-Instruct}} \\ \hline
UnKE & 93.56 / 78.09 & 93.61 / 79.26 & 91.42 / 71.73 & 93.33 / 78.42 & 98.34 / 93.38 & 37.87 \\
UnKE+Any & \textbf{99.54} / 72.96 & 99.48 / 78.24 & \textbf{99.29} / 70.92 & 99.62 / 77.48 & \textbf{99.79} / 91.63 & 38.89 \\
\rowcolor[HTML]{ECF4FF} 
\textbf{UnKE+OUR} & 99.19 / \textbf{79.94} & \textbf{99.89 / 82.27} & 99.01 / \textbf{75.46} & \textbf{99.80 / 81.57} & 99.37 / \textbf{96.71} & \textbf{42.7} \\
MEMIT & 25.57 / 22.88 & 32.67 / 30.75 & 14.51 / 12.37 & 30.49 / 28.65 & 76.21 / 74.25 & 22.56 \\
MEMIT+Any & 78.92 / 67.96 & 91.28 / 81.77 & 87.72 / 74.27 & 90.36 / \textbf{80.51} & 95.86 / 92.5 & 41.67 \\
\rowcolor[HTML]{ECF4FF} 
\textbf{MEMIT+OUR} & \textbf{91.60 / 80.30} & \textbf{92.36 / 82.02} & \textbf{89.45 / 77.31} & \textbf{92.62} / 78.56 & \textbf{98.31 / 95.56} & \textbf{46.93} \\
AlphaEdit & 21.29 / 20.24 & 28.62 / 27.99 & 11.36 / 10.24 & 26.59 / 25.92 & 73.92 / 72.96 & 20.71 \\
AlphaEdit+Any & 62.98 / 52.44 & 79.68 / 69.13 & 71.34 / 56.67 & 78.79 / 67.77 & 91.71 / 87.41 & 42.86 \\
\rowcolor[HTML]{ECF4FF} 
\textbf{AlphaEdit+OUR} & \textbf{75.63 / 61.69} & \textbf{81.38 / 73.66} & \textbf{79.38 / 60.10} & \textbf{80.27 / 69.10} & \textbf{93.01 / 88.12} & \textbf{43.73} \\ \hline
\multicolumn{7}{c}{\textbf{Based on Qwen2.5-7B-Instruct}} \\ \hline
UnKE & 91.72 / 56.79 & 91.41 / \textbf{68.47} & 87.75 / \textbf{56.34} & 91.01 / 51.97 & 96.97 / 83.85 & 38.12 \\
UnKE+Any & 96.33 / 47.42 & 97.42 / 53.18 & 96.33 / 35.31 & 97.31 / 50.93 & 98.83 / 80.87 & 33.43 \\
\rowcolor[HTML]{ECF4FF} 
\textbf{UnKE+OUR} & \textbf{99.54 / 66.13} & \textbf{99.46} / 67.68 & \textbf{99.26} / 54.32 & \textbf{99.44 / 66.12} & \textbf{99.71 / 88.24} & \textbf{37.88} \\
MEMIT & 45.07 / 40.81 & 40.73 / 36.75 & 19.59 / 15.87 & 38.04 / 34.07 & 78.03 / 76.50 & 24.75 \\
MEMIT+Any & 86.59 / 64.61 & 91.84 / 72.49 & 88.63 / 61.24 & 91.52 / 71.12 & 96.85 / 89.85 & 41.41 \\
\rowcolor[HTML]{ECF4FF} 
\textbf{MEMIT+OUR} & \textbf{95.96 / 77.49} & \textbf{96.92 / 79.55} & \textbf{95.56 / 70.85} & \textbf{96.77 / 78.56} & \textbf{99.71 / 93.27} & \textbf{43.66} \\
AlphaEdit & 49.71 / 45.21 & 45.42 / 41.06 & 24.63 / 19.85 & 42.77 / 38.26 & 80.48 / 78.38 & 25.37 \\
AlphaEdit+Any & 97.77 / 83.09 & 98.20 / 84.18 & 97.40 / 77.38 & 98.14 / 83.40 & 99.08 / 94.51 & 41.58 \\
\rowcolor[HTML]{ECF4FF} 
\textbf{AlphaEdit+OUR} & \textbf{99.28 / 86.16} & \textbf{99.33 / 87.16} & \textbf{99.01 / 81.35} & \textbf{99.29 / 86.46} & \textbf{99.7 / 94.6} & \textbf{43.48} \\ \hline
\multicolumn{7}{c}{\textbf{Based on Qwen3-8B}} \\ \hline
MEMIT & 19.46 / 19.20 & 36.21 / 35.74 & 12.55 / 11.31 & 34.01 / 33.49 & 66.96 / 66.51 & 29.18 \\
MEMIT+Any & 31.29 / 25.88 & 53.48 / 47.21 & 34.41 / 25.81 & 55.47 / 44.81 & 76.78 / 72.95 & 32.58 \\
\rowcolor[HTML]{ECF4FF} 
\textbf{MEMIT+OUR} & \textbf{44.52 / 33.2} & \textbf{63.28 / 52.22} & \textbf{48.39 / 33.88} & \textbf{61.64 / 50.11} & \textbf{82.56 / 77.1} & \textbf{34.31} \\
AlphaEdit & 23.47 / 22.25 & 32.23 / 30.99 & 12.72 / 11.06 & 30.01 / 28.83 & 62.54 / 62.88 & 25.87 \\
AlphaEdit+Any & 40.88 / 31.4 & 58.27 / 49.85 & 41.88 / 30.18 & 56.56 / 47.73 & 79.19 / 75.3 & 33.01 \\
\rowcolor[HTML]{ECF4FF} 
\textbf{AlphaEdit+OUR} & \textbf{68.16 / 48.42} & \textbf{76.75 / 61.23} & \textbf{67.44 / 46.34} & \textbf{75.85 / 59.64} & \textbf{88.58 / 81.69} & \textbf{36.11} \\ \hline
\end{tabular}
\label{all}
\end{table}

\textbf{DynEdit consistently improves performance across all editors.}
When combined with MEMIT, UnKE, and AlphaEdit, DynEdit uniformly outperforms their AnyEdit counterparts on both lexical and semantic metrics. The gains are particularly evident in paraphrased settings (second numbers), indicating stronger robustness to input variation.

\textbf{The improvements generalize across base models.}
DynEdit achieves consistent gains on both Llama3-8B and Qwen2.5-7B, and remains effective on the stronger Qwen3-8B model, where it significantly improves over AnyEdit despite the increased baseline difficulty.

\textbf{DynEdit yields larger gains for editors that are less effective in long-sequence settings.}
The improvements are more pronounced when applied to editors such as MEMIT and AlphaEdit under long-target scenarios, where fixed or uniform editing strategies struggle to capture complex dependencies. This suggests that entropy-guided allocation helps compensate for limited long-sequence editing capacity by focusing optimization on critical tokens.

Overall, these results demonstrate that DynEdit is a plug-and-play enhancement that consistently improves different editing frameworks across models and evaluation settings.

\subsection{Time complexity analysis}
\label{efficiency}

We further analyze the computational cost of DynEdit compared with representative editing baselines. Since DynEdit follows the same parameter-update procedure as its backbone editor, its main additional cost comes from token-level entropy estimation and adaptive sequential scheduling. The entropy map is computed once before editing, and the extra memory required for storing token-level entropy and weights is negligible compared with the memory footprint of the base LLM and the edited modules.

Let $M$ denote the target length, $w$ the window size, and $N$ the number of editing windows. Fixed-window methods such as AnyEdit use a deterministic schedule, while DynEdit may introduce denser windows in high-entropy regions. Therefore, the time complexity of DynEdit is mainly proportional to the number of dynamically selected windows:
\begin{equation}
    \mathcal{O}\!\left(N_{\mathrm{Dyn}} \cdot C_{\mathrm{edit}}\right),
\end{equation}
where $C_{\mathrm{edit}}$ denotes the cost of optimizing one local edit. Since DynEdit keeps the same edited layers and update modules as AnyEdit, its memory complexity remains essentially unchanged. The adaptive scheduling increases runtime moderately, but does not introduce additional large-scale parameters or auxiliary models.

\begin{figure*}[h]
    \centering
    \begin{minipage}[t]{0.48\textwidth}
        \centering
        \vspace{0pt}
        \includegraphics[width=0.92\linewidth]{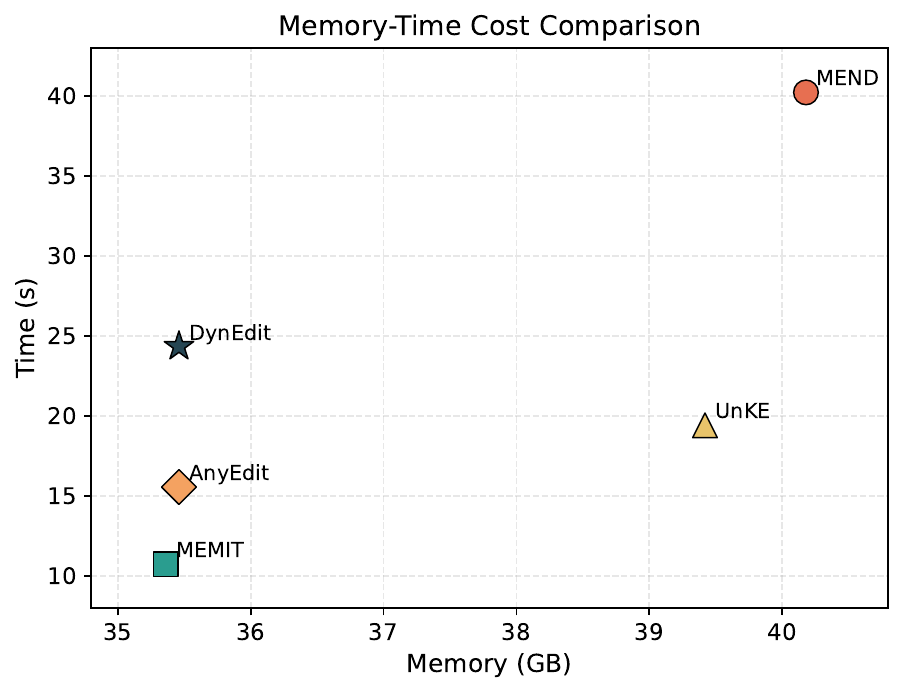}
        \caption{
        Memory-time cost comparison.
        }
        \label{fig:efficiency}
    \end{minipage}
    \hfill
    \begin{minipage}[t]{0.48\textwidth}
        \centering
        \captionof{table}{
        Efficiency comparison on long-sequence editing.
        }
        \vspace{0pt}
        \small
        \setlength{\tabcolsep}{3.5pt}
        \renewcommand{\arraystretch}{1.05}
        \resizebox{\linewidth}{!}{
        \begin{tabular}{lccc}
        \toprule
        \textbf{Method} & \textbf{Memory (GB)} & \textbf{Time (s)} & \textbf{Rel. Time} \\
        \midrule
        MEND    & 40.18 & 40.23 & 2.58$\times$ \\
        MEMIT   & 35.36 & 10.75 & 0.69$\times$ \\
        UnKE    & 39.42 & 19.42 & 1.25$\times$ \\
        AnyEdit & 35.46 & 15.57 & 1.00$\times$ \\
        DynEdit & 35.46 & 24.36 & 1.56$\times$ \\
        \bottomrule
        \end{tabular}
        }
        \label{tab:efficiency}
    \end{minipage}
\end{figure*}

Table~\ref{tab:efficiency} reports the empirical memory and time cost. DynEdit achieves the same peak memory usage as AnyEdit, while introducing a moderate runtime overhead due to entropy-guided dynamic sliding and weighted suffix optimization. Compared with MEND and UnKE, DynEdit remains more memory-efficient, and its runtime stays within a practical range for long-sequence editing.

\subsection{Batch Editing performance}
\label{batch}
We further evaluate DynEdit in the \emph{batch editing} setting, where multiple knowledge updates are applied sequentially to the same model. This scenario is particularly challenging, as accumulated edits often introduce interference, leading to performance degradation and reduced generalization.
\begin{figure}[h]
  \centering
  \includegraphics[width=\textwidth]{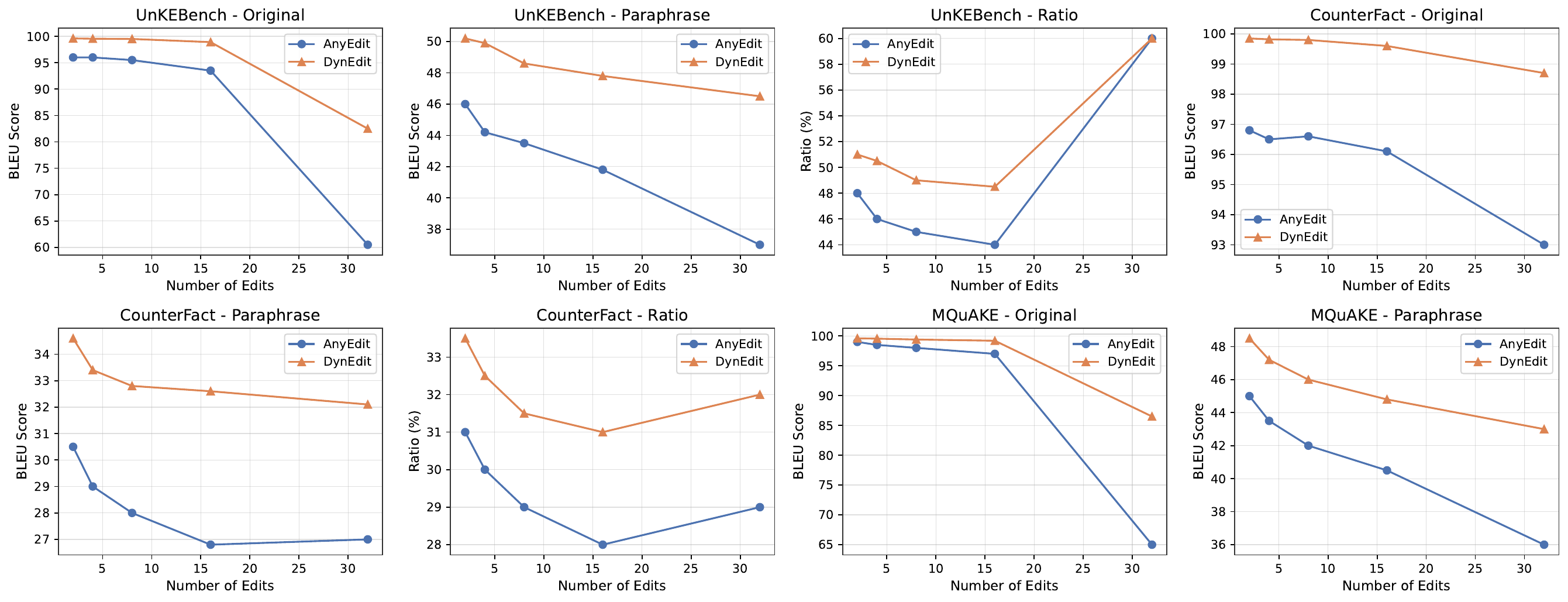}
  \caption{Batch editing performance of DynEdit and AnyEdit on three benchmarks.}\label{fig:batch}
\end{figure}

Figure~\ref{fig:batch} compares DynEdit with strong baselines across increasing numbers of edits. We observe that while all methods suffer from degradation as the number of edits grows, DynEdit consistently maintains higher performance on both original and paraphrased queries. The advantage becomes more pronounced in high-edit regimes (e.g., 16$\sim$32 edits), where competing methods exhibit sharp drops.

More importantly, we analyze the \emph{relative performance preservation} by measuring the ratio between paraphrase and original BLEU scores. This metric reflects the robustness of edited knowledge under linguistic variation. As shown in Figure~\ref{fig:batch} (right), DynEdit achieves consistently higher ratios compared to baselines, indicating stronger generalization beyond the original surface forms. In contrast, existing methods show noticeable declines, suggesting that their edits tend to overfit to specific phrasings.

These results demonstrate that DynEdit not only improves absolute editing accuracy, but also enhances robustness under repeated edits. We attribute this to its entropy-guided design: dynamic window scheduling reduces unnecessary updates on low-information regions, while entropy-weighted optimization mitigates gradient conflict and prevents critical knowledge from being diluted. As a result, DynEdit produces more stable and semantically consistent edits in the batch setting.

\subsection{Hyperparameter analysis}
\label{hyper}
We further analyze the sensitivity of DynEdit to two key allocation hyperparameters: the base window size $w$ and the dense stride ratio $\Delta_{\mathrm{dense}}/w$. The base window size controls the local context range used in each editing step, while the dense stride ratio determines how frequently high-entropy regions are revisited during dynamic scheduling. We report the results on UnKE-Paraphrase for both Llama-3-8B and Qwen-2.5-7B.

\begin{figure}[h]
  \centering
  \includegraphics[width=\textwidth]{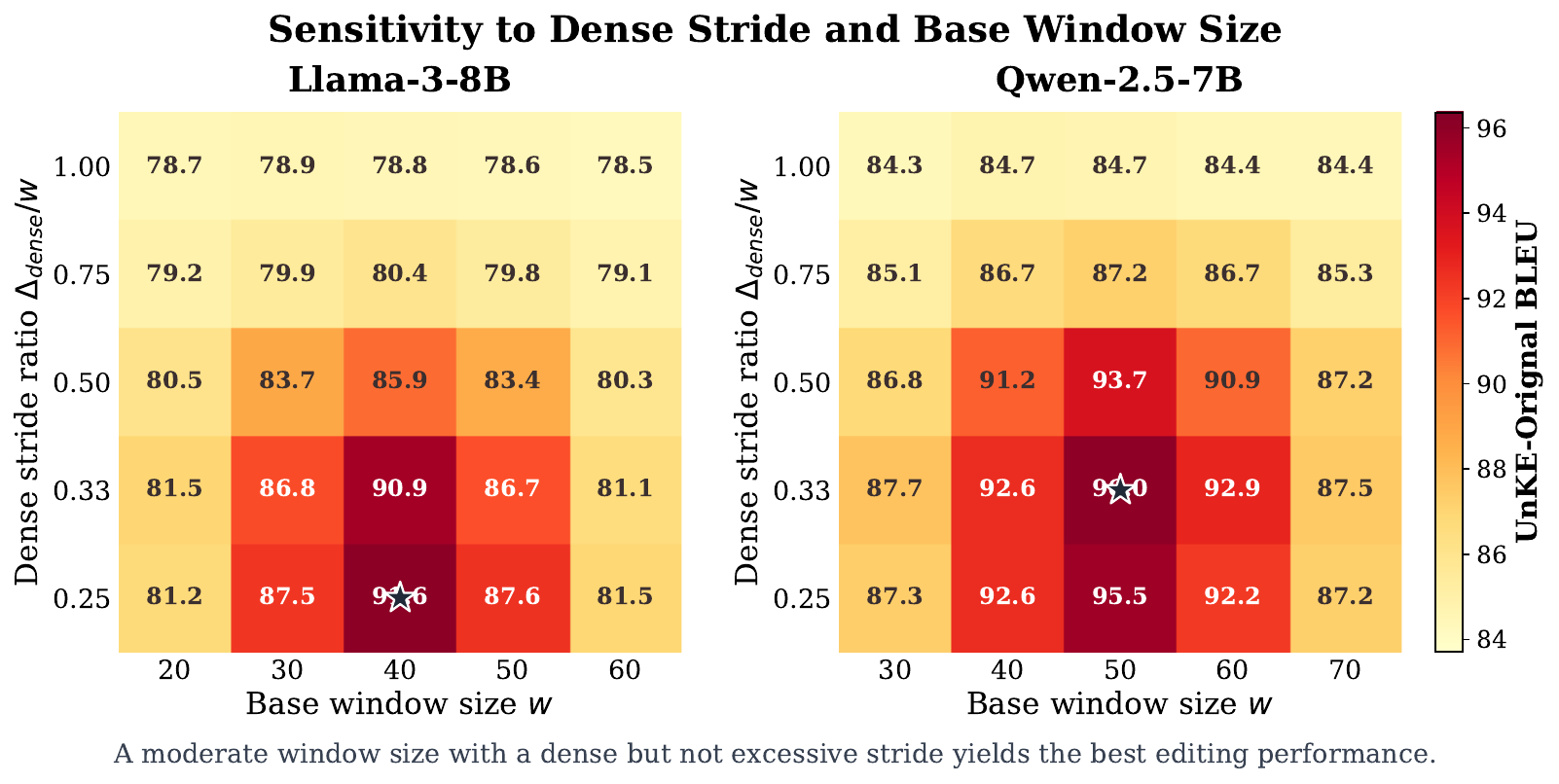}
  \caption{Hyperparameter analysis of DynEdit on UnKE-Original. The heatmaps show the effect of base window size $w$ and dense stride ratio $\Delta_{\mathrm{dense}}/w$ on editing performance.}
  \label{fig:hyperparameter}
\end{figure}

As shown in Figure~\ref{fig:hyperparameter}, DynEdit achieves the best performance under moderate window sizes and dense but not excessive strides. When the stride ratio is too large, the editing trajectory becomes close to fixed-window editing, leading to insufficient coverage of difficult high-entropy regions. In contrast, an overly small stride introduces redundant local optimization, which brings limited additional benefit and may slightly reduce stability. Similarly, very small windows provide insufficient context for long-form targets, while overly large windows weaken the locality of entropy-guided allocation. Overall, DynEdit remains robust across a broad range of settings, and its best performance is obtained when window size and dense stride are balanced to provide focused yet stable optimization.

\subsection{Case Studies}
\label{case}
We present the case studies from UnKE. The two cases show that AnyEdit can still fail when the edited fact contains salient but position-sensitive tokens. In the first case, the key evidence lies in the specific comic titles, where AnyEdit replaces the target titles with plausible but incorrect alternatives. In the second case, the edited birthplace is the central factual token, yet AnyEdit consistently generates an incorrect location and further expands it into redundant background descriptions. These failures indicate that uniform editing over long target sequences may underemphasize locally critical tokens, causing the model to preserve fluent but factually misaligned responses.

\begin{center}
\scriptsize
\setlength{\fboxsep}{2pt}

\begin{tcolorbox}[
    width=\textwidth,
    colback=white,
    colframe=casegray,
    arc=1mm,
    boxrule=0.5pt,
    left=2pt,right=2pt,top=2pt,bottom=2pt,
    title=\textbf{Case Study on UnKE},
    coltitle=white,
    colbacktitle=casegray,
    fonttitle=\bfseries\scriptsize
]

\textbf{Case 1. Cartoonist Achievements.}
\textbf{Q:} What are some notable achievements and characteristics of Marco Ameglio as a cartoonist?
\textbf{Paraphrase:} What are some significant accomplishments and distinguishing qualities of Marco Ameglio in his career as a cartoonist?

\textbf{Edit Target:}
Marco Ameglio is a talented cartoonist who has been creating comics for over a decade. He has published several successful comic book series, including \correct{``The Adventures of Captain Comet''} and \correct{``The Misadventures of Max and Molly.''} Ameglio has also created \correct{custom comics for advertising agencies and non-profit organizations}. His unique style and attention to detail have earned him a loyal following.

\vspace{0.25em}
\begin{minipage}[t]{0.49\textwidth}
\begin{tcolorbox}[
    colback=lightred,
    colframe=casered,
    arc=1mm,
    boxrule=0.45pt,
    left=2pt,right=2pt,top=2pt,bottom=2pt,
    title=\textbf{AnyEdit},
    coltitle=white,
    colbacktitle=casered,
    fonttitle=\bfseries\scriptsize
]
\textbf{Original:}
Marco Ameglio has published several successful comic book series, including \wrong{``The Adventures of Captain Cosmos''} and \wrong{``The Secret Life of Pets''}. His comics have been praised for engaging storylines and beautiful artwork. Some notable achievements include \wrong{publishing 10 comic book series, winning awards, and collaborating with other artists}.

\textbf{Paraphrase:}
Marco Ameglio has published several successful comic book series, including \wrong{``The Adventures of Captain Cosmos''} and \wrong{``The Secret Life of Pets''}. Some notable accomplishments include \wrong{publishing several comic book series, winning awards, and collaborating with other artists}.
\end{tcolorbox}
\end{minipage}
\hfill
\begin{minipage}[t]{0.49\textwidth}
\begin{tcolorbox}[
    colback=lightblue,
    colframe=caseblue,
    arc=1mm,
    boxrule=0.45pt,
    left=2pt,right=2pt,top=2pt,bottom=2pt,
    title=\textbf{DynEdit},
    coltitle=white,
    colbacktitle=caseblue,
    fonttitle=\bfseries\scriptsize
]
\textbf{Original:}
Marco Ameglio has published several successful comic book series, including \correct{``The Adventures of Captain Comet''} and \correct{``The Misadventures of Max and Molly.''} He has also created \correct{custom comics for advertising agencies and non-profit organizations}. His unique style and attention to detail have earned him a loyal following.

\textbf{Paraphrase:}
Marco Ameglio has published several successful comic book series, including \correct{``The Adventures of Captain Comet''} and \correct{``The Misadventures of Max and Molly.''} He has also created \correct{comics for advertising agencies and non-profit organizations}. His unique style and attention to detail have earned him a loyal following.
\end{tcolorbox}
\end{minipage}

\vspace{0.35em}
\textbf{Case 2. Birthplace Editing.}
\textbf{Q:} Where was the contemporary artist Danie Mellor born?
\textbf{Paraphrase:} What is the birthplace of the modern artist Danie Mellor?

\textbf{Edit Target:}
Danie Mellor was born in \correct{Chertsey}, a small town in \correct{Surrey, England}. His official biography states that Mellor was born in \correct{1971 in Chertsey} and later migrated to Australia with his family in \correct{1973}. Therefore, Danie Mellor was indeed born in \correct{Chertsey}.

\vspace{0.25em}
\begin{minipage}[t]{0.49\textwidth}
\begin{tcolorbox}[
    colback=lightred,
    colframe=casered,
    arc=1mm,
    boxrule=0.45pt,
    left=2pt,right=2pt,top=2pt,bottom=2pt,
    title=\textbf{AnyEdit},
    coltitle=white,
    colbacktitle=casered,
    fonttitle=\bfseries\scriptsize
]
\textbf{Original:}
Danie Mellor was born in \wrong{Kuala Lumpur, Malaysia}. According to his official website, Mellor was born in \wrong{1971 in Kuala Lumpur}. Mellor's \wrong{Malaysian heritage} has had a significant influence on his work.

\textbf{Paraphrase:}
Danie Mellor was born in \wrong{Kuala Lumpur, Malaysia}. He grew up in Brisbane, Queensland. Mellor's \wrong{mixed heritage} influenced his work, which often incorporates elements of Aboriginal and Asian cultures.
\end{tcolorbox}
\end{minipage}
\hfill
\begin{minipage}[t]{0.49\textwidth}
\begin{tcolorbox}[
    colback=lightblue,
    colframe=caseblue,
    arc=1mm,
    boxrule=0.45pt,
    left=2pt,right=2pt,top=2pt,bottom=2pt,
    title=\textbf{DynEdit},
    coltitle=white,
    colbacktitle=caseblue,
    fonttitle=\bfseries\scriptsize
]
\textbf{Original:}
Danie Mellor was born in \correct{Chertsey}, a small town in \correct{Surrey, England}. According to his official website, Mellor was born in \correct{1971 in Chertsey} and later moved to Australia with his family.

\textbf{Paraphrase:}
Danie Mellor was born in \correct{Chertsey}, a small town in \correct{Surrey, England}. His biography states that he was \correct{``born in Chertsey, England, and migrated to Australia with his family in 1973.''}
\end{tcolorbox}
\end{minipage}

\end{tcolorbox}
\label{fig:case_study}
\end{center}

In contrast, DynEdit better preserves the edited knowledge under both original and paraphrased prompts. By using entropy-aware weighting, DynEdit assigns stronger optimization pressure to uncertain and informative token positions. As a result, DynEdit produces responses that are more tightly aligned with the edit target and less prone to verbose hallucinated continuations.

\subsection{Editing Allocation Visualization.}

\begin{figure}[h]
  \centering
  \includegraphics[width=\textwidth]{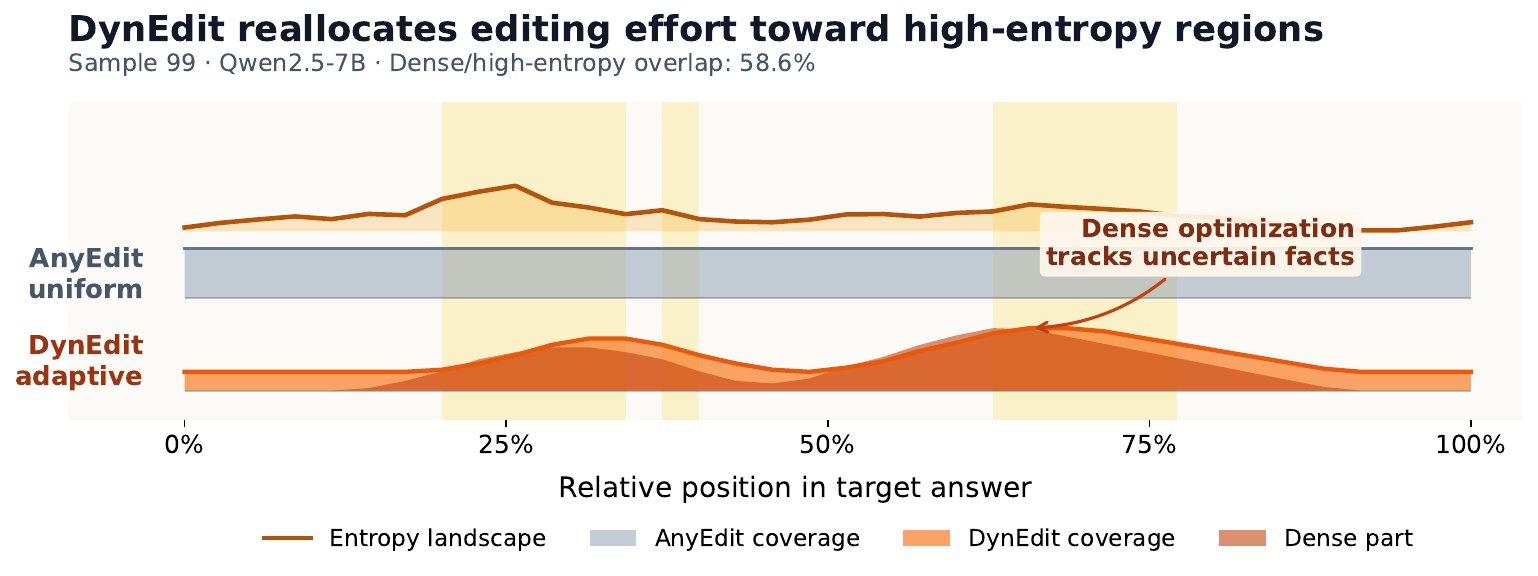}
  \caption{Token-level analysis of entropy, editing difficulty, and semantic role. High-entropy tokens are both harder to edit and more likely to carry core semantic information.}
  \label{fig:dynedit_allocation}
\end{figure}

Fig.~\ref{fig:dynedit_allocation} visualizes how editing effort is distributed along a target answer. The entropy curve indicates the token-level uncertainty of the pre-edit model, while the lower bands show the optimization coverage of AnyEdit and DynEdit. AnyEdit follows a nearly uniform window allocation, whereas DynEdit adaptively increases coverage in high-entropy regions. This suggests that DynEdit performs more focused optimization on locally difficult and factually important positions, rather than uniformly editing the entire target sequence.

\section{Limitations}
\label{sec:limitations}

Although DynEdit achieves consistent improvements in long-sequence knowledge editing, several limitations remain. First, DynEdit relies on token-level entropy estimated from the pre-edit model as a proxy for editing difficulty and semantic importance. While effective in our experiments, entropy may not always perfectly reflect the true importance of a token, especially in highly structured or domain-specific generation tasks.

Second, the current framework is designed for sequential text generation and has been evaluated mainly on long-form factual editing benchmarks. Its effectiveness in more complex settings, such as multi-modal editing, multi-turn interactive editing, or reasoning-intensive generation, remains underexplored. n future work, we plan to incorporate graph pre-training techniques \cite{11601110, yantowards} to represent entities, relations, and higher-order dependencies, thereby extending long-sequence knowledge editing to more complex multi-relation reasoning scenarios. Building on existing work on multimodal safety enhancement, \cite{fu2026diagnosing}, we will also investigate how long-sequence editing can be extended to safety-oriented applications.

In addition, DynEdit introduces additional scheduling and weighting operations during optimization. Although the overhead is moderate, the editing cost still increases compared with fixed-window methods. Further improving efficiency while maintaining editing quality is an important direction for future work.

More broadly, we hope this work can encourage future research on adaptive knowledge editing strategies that dynamically allocate editing capacity according to the internal structure and uncertainty of model knowledge.

\section{Broader Impact}
\label{Broader}

This work studies long-sequence knowledge editing for large language models. Efficient and reliable knowledge editing can provide positive societal benefits by enabling models to correct outdated, inaccurate, or conflicting information without expensive retraining. Improving long-form editing may further enhance the factual consistency and controllability of deployed language models in applications such as education, question answering, and information services.

At the same time, knowledge editing techniques may also introduce potential risks if misused. For example, malicious users could intentionally modify model behavior to inject misleading information, manipulate generated content, or bypass safety constraints. As our method improves the effectiveness of long-sequence editing, it could potentially lower the cost of such manipulations.

Our work is intended for controlled and transparent model adaptation in research settings. We hope this research can contribute to a better understanding of reliable and interpretable knowledge updating mechanisms, while encouraging future work on safety evaluation, misuse detection, and robust editing constraints for large language models.


\clearpage
\input{checklist.tex}

\end{document}

%% file: checklist.tex
\section*{NeurIPS Paper Checklist}

\begin{enumerate}

\item {\bf Claims}
    \item[] Question: Do the main claims made in the abstract and introduction accurately reflect the paper's contributions and scope?
    \item[] Answer: \answerYes{} 
    \item[] Justification: Our abstract and introduction clearly state the arguments and contributions of the paper. See Section \ref{intro}.

\item {\bf Limitations}
    \item[] Question: Does the paper discuss the limitations of the work performed by the authors?
    \item[] Answer: \answerYes{} 
    \item[] Justification: Section Appendix \ref{sec:limitations}.

\item {\bf Theory assumptions and proofs}
    \item[] Question: For each theoretical result, does the paper provide the full set of assumptions and a complete (and correct) proof?
    \item[] Answer: \answerYes{} 
    \item[] Justification: Section Appendix \ref{Proof}.

    \item {\bf Experimental result reproducibility}
    \item[] Question: Does the paper fully disclose all the information needed to reproduce the main experimental results of the paper to the extent that it affects the main claims and/or conclusions of the paper (regardless of whether the code and data are provided or not)?
    \item[] Answer: \answerYes{} 
    \item[] Justification: Section Appendix \ref{subsec:experimental_setup}.

\item {\bf Open access to data and code}
    \item[] Question: Does the paper provide open access to the data and code, with sufficient instructions to faithfully reproduce the main experimental results, as described in supplemental material?
    \item[] Answer: \answerYes{} 
    \item[] Justification: We provide it in supplemental material.

\item {\bf Experimental setting/details}
    \item[] Question: Does the paper specify all the training and test details (e.g., data splits, hyperparameters, how they were chosen, type of optimizer) necessary to understand the results?
    \item[] Answer: \answerYes{} 
    \item[] Justification: Section Appendix \ref{subsec:experimental_setup}.

\item {\bf Experiment statistical significance}
    \item[] Question: Does the paper report error bars suitably and correctly defined or other appropriate information about the statistical significance of the experiments?
    \item[] Answer: \answerYes{} 
    \item[] Justification: Section 5 and Appendix C.3. Error bars correspond to one standard deviation across runs.

\item {\bf Experiments compute resources}
    \item[] Question: For each experiment, does the paper provide sufficient information on the computer resources (type of compute workers, memory, time of execution) needed to reproduce the experiments?
    \item[] Answer: \answerYes{} 
    \item[] Justification: Section Appendix \ref{subsec:experimental_setup}.
    
\item {\bf Code of ethics}
    \item[] Question: Does the research conducted in the paper conform, in every respect, with the NeurIPS Code of Ethics \url{https://neurips.cc/public/EthicsGuidelines}?
    \item[] Answer: \answerYes{} 
    \item[] Justification: We strictly adhere to the NeurIPS Code of Ethics.

\item {\bf Broader impacts}
    \item[] Question: Does the paper discuss both potential positive societal impacts and negative societal impacts of the work performed?
    \item[] Answer: \answerYes{} 
    \item[] Justification: Section Appendix \ref{Broader}.
    
\item {\bf Safeguards}
    \item[] Question: Does the paper describe safeguards that have been put in place for responsible release of data or models that have a high risk for misuse (e.g., pre-trained language models, image generators, or scraped datasets)?
    \item[] Answer: \answerNA{} 
    \item[] Justification: The paper poses no such risks.

\item {\bf Licenses for existing assets}
    \item[] Question: Are the creators or original owners of assets (e.g., code, data, models), used in the paper, properly credited and are the license and terms of use explicitly mentioned and properly respected?
    \item[] Answer: \answerYes{} 
    \item[] Justification: Section Reference.

\item {\bf New assets}
    \item[] Question: Are new assets introduced in the paper well documented and is the documentation provided alongside the assets?
    \item[] Answer: \answerNA{} 
    \item[] Justification: We do not involve any risk datasets or models.

\item {\bf Crowdsourcing and research with human subjects}
    \item[] Question: For crowdsourcing experiments and research with human subjects, does the paper include the full text of instructions given to participants and screenshots, if applicable, as well as details about compensation (if any)? 
    \item[] Answer: \answerNA{}  
    \item[] Justification: This item is not involved.

\item {\bf Institutional review board (IRB) approvals or equivalent for research with human subjects}
    \item[] Question: Does the paper describe potential risks incurred by study participants, whether such risks were disclosed to the subjects, and whether Institutional Review Board (IRB) approvals (or an equivalent approval/review based on the requirements of your country or institution) were obtained?
    \item[] Answer: \answerNA{} 
    \item[] Justification: This item is not involved.

\item {\bf Declaration of LLM usage}
    \item[] Question: Does the paper describe the usage of LLMs if it is an important, original, or non-standard component of the core methods in this research? Note that if the LLM is used only for writing, editing, or formatting purposes and does \emph{not} impact the core methodology, scientific rigor, or originality of the research, declaration is not required.
    \item[] Answer: \answerNA{} 
    \item[] Justification: This item is not involved.

\end{enumerate}